\documentclass[10pt,journal,compsoc,twocolumn]{IEEEtran}
\ifCLASSOPTIONcompsoc
  \usepackage[nocompress]{cite}
\else
  \usepackage{cite}
\fi
\usepackage{graphicx,multicol,multirow}

\ifCLASSINFOpdf
\else
\fi
\usepackage{amsmath}
\usepackage{amssymb}
\begin{document}
%
\title{Alternating Iteratively Reweighted Minimization Algorithms for Low-Rank Matrix Factorization}
%
%
%
%
\author{Paris V. Giampouras,
        Athanasios A. Rontogiannis and Konstantinos D. Koutroumbas
\IEEEcompsocitemizethanks{\IEEEcompsocthanksitem The authors are with the Institute for Astronomy, Astrophysics, Space Applications and Remote Sensing at the National Observatory of Athens, Penteli, 15236, Greece.\protect}
}
%
%
\markboth{Journal of \LaTeX\ Class Files,~Vol.~14, No.~8, August~2015}%
{Shell \MakeLowercase{\textit{et al.}}: Bare Demo of IEEEtran.cls for Computer Society Journals}
%
\IEEEtitleabstractindextext{%
\begin{abstract}
Nowadays, the availability of large-scale data in disparate application domains urges the deployment of sophisticated tools for extracting valuable knowledge out of this huge bulk of information. In that vein, low-rank representations (LRRs) which seek low-dimensional  embeddings of data have naturally appeared. In an effort to reduce computational complexity and improve estimation performance, LRR has been viewed via a matrix factorization (MF) perspective. Recently, low-rank MF (LRMF) approaches have been proposed for tackling the inherent weakness of MF i.e., the unawareness of the dimension of the low-dimensional space where data reside. Herein, inspired by the merits of iterative reweighted schemes for rank minimization, we come up with a generic low-rank promoting regularization function. Then, focusing on a specific instance of it, we propose a regularizer that imposes column-sparsity jointly on the two matrix factors that result from MF, thus promoting low-rankness on the optimization problem. The problems of denoising, matrix completion and non-negative matrix factorization (NMF) are redefined according to the new LRMF formulation and solved via efficient Newton-type algorithms with proven theoretical guarantees as to their convergence and rates of convergence to stationary points. The effectiveness of the proposed algorithms is verified in diverse simulated and real data experiments. 
\end{abstract}
}
\maketitle
\IEEEdisplaynontitleabstractindextext
%
\IEEEpeerreviewmaketitle
\IEEEraisesectionheading{\section{Introduction}\label{sec:introduction}}
Low-rank representation (LRR) of data has recently attracted great interest since it appears in a wide spectrum of research fields and applications, such as signal processing, machine learning, quantum tomography, etc, \cite{theodoridis2015machine}. LRR shares similar characteristics with sparse representation and hence is in principle formulated as a NP-hard problem, \cite{fazel2002}. Convex relaxations have played a remarkable role in the course of making the problem tractable. In that respect, the nuclear norm has been extensively applied offering favorable results, optimal recovery performance, as well as a solid theoretical understanding, \cite{recht}. However, in the case of high-dimensional and large-scale datasets, conventional convex LRR approaches are confronted with inherent limitations related to their high computational complexity, \cite{hastie2015matrix}. 

To overcome these limitations matrix factorization (MF) methods have been introduced lately. MF gives rise to non-convex optimization problems and hence its theoretical understanding is a much more challenging task. Notably, a great effort has been recently devoted towards deriving a comprehensive theoretical framework of MF with the goal to reach to optimal recovery guarantees, \cite{Sun,ge2017,zhu}. MF presents significant computational merits by reducing the size of the emerging optimization problems. Thus, it leads to optimization algorithms of lower computational complexity as compared to relevant convex approaches. In addition, MF lies at the heart of a variety of problems dealing with the task of finding low-rank embeddings. In that respect, ubiquitous problems such as clustering, \cite{pompili}, blind source separation, matrix completion, \cite{wen} etc. have been seen in literature through the lens of MF. MF entails the use of two matrix factors with a fixed number of columns, which, in the most favorable case, coincides with the rank of the sought matrix.   However, the rank of the matrix, which is usually much less than its dimensions, is unknown a priori.

In light of this, a widespread approach is based on the following premise: overstate the number of columns of the matrix factors and then penalize their rank by using appropriate low-rank promoting regularizers. Along those lines, various regularizers have been recently proposed. Amongst them the most popular one is the variational characterization of the nuclear norm (proven to be a tight upper-bound of it) defined as the sum of the squared Frobenious norms of the factors \cite{srebro2005rank}. More recently, generalized versions of this approach have come to the scene. In that respect, in \cite{shang}, tight upper-bounds of the low-rank promoting Schatten-$p$ norms were presented under a general framework. In \cite{haeffele2014structured}, an alternative approach for promoting low-rankness via non-convex MF was described. The novelty of that approach comes from the incorporation of additional constraints on the matrix factors giving thus rise to an interesting low-rank structured MF framework. In \cite{hastie2015matrix}, a fast algorithm based on the above-mentioned variational characterization of the nuclear norm is presented. The derived algorithm is amenable to handling incomplete big-data, contrary to conventional convex and other non-MF based approaches. It should be noted that common characteristic of all state-of-the-art methods is the following: although the rank of the product of the matrix factors may decrease as a result of the penalization process, {\it the  number of columns of the matrix factors (which has initially been overstated) remains fixed throughout the execution of the minimization algorithms}. Hence, the per iteration complexity remains unaltered, albeit the rank of the matrix factors may potentially decrease gradually to a large degree as the algorithms evolve. 

With the current work we capitalize on the latter (possibly undesirable in large-scale data applications) issue and propose a novel generic formulation for non-convex low-rank MF. To this end, recent ideas stemming from iterative reweighted approaches for low-rank matrix estimation, proposed in \cite{fornasier2011low,mohan2012iterative} as efficient alternatives for nuclear norm minimization, are now extended to the MF framework. This way, we come up with a novel alternating reweighted scheme for low-rank promotion in MF problems. As is shown, the recent low-rank MF schemes proposed in \cite{shang} can be cast as special occasions of the proposed formulation by suitably selecting the reweighting matrices applied on the matrix factors. Going one step further, we propose the  selection of a common reweighting matrix that couples the matrix factors and leads to a joint column sparsity promoting regularization term, \cite{ssp2016,spars2017}. In doing so, {\it low-rank promotion now reduces to the task of jointly annihilating columns of the matrix factors}. Interestingly, {\it this way the computational complexity of the derived algorithms decreases progressively, since the size of the estimated matrix factors is reduced as the algorithms evolve}. 

In an effort to better highlight the efficiency and ubiquity of the proposed low-rank MF formulation, we address three popular problems in the machine learning literature, namely denoising, matrix completion and non-negative matrix factorization. These problems are accordingly formulated in Section 2. By exploiting novel optimization concepts, \cite{hong2016unified}, we appropriately minimize the arising non-smooth and non-separable cost functions. In this vein, novel second-order Newton-type algorithms are then devised in Section 3 with the goal to effectively exploit inherent characteristics of the emerging optimization problems. Convergence analysis of the algorithms at stationary points and their rates of convergence are given in Section 4. In Section 5, the merits of the resulting algorithms in terms of estimation performance and computational complexity, compared to relevant state-of-art algorithms, are illustrated on simulated and real data experiments. In order to test the effectiveness of the proposed algorithms on real applications involving large-scale data, the problems of hyperspectral image denoising, matrix completion in movies recommender systems and music signal decomposition are employed.
\section{Low-rank matrix factorization}\label{sec:introduction}
Low-rank matrix estimation per se has been addressed by a wealth of different approaches, lending itself to disparate applications. Focusing on the task of recovering low-rank matrices from linear measurements, we come up with the ubiquitous affine rank minimization problem, \cite{recht}, which is formulated as follows,
\begin{align}
\mathrm{min}\left[ \mathrm{rank}(\mathbf{X}) \right] \;\;\;\ s.t \;\;\;\;\ \mathbf{\mathcal{A}}(\mathbf{X})= \mathbf{b}, \label{pb_1:af_rank_minim}
\end{align}
where $\mathbf{\mathcal{A}}$ denotes the linear operator that maps $\mathbf{X}\in\mathcal{R}^{m\times n}$ to $\mathbf{b}\in\mathcal{R}^l$. Problem (\ref{pb_1:af_rank_minim}) is tantamount to solving the $\ell_0$ minimization problem on the singular values of $\mathbf{X}$ and hence is NP-hard. To this end various relaxation schemes have come to the scene in literature, many of which are based on the Schatten-$p$ norm\cite{Nie,Lu2014}. The Schatten-$p$ norm is defined as, 
\begin{align}
\|\mathbf{X}\|_{\mathcal{S}_p} = \|\boldsymbol{\sigma}(\mathbf{X})\|_p,
\end{align} 
where $\boldsymbol{\sigma}(\mathbf{X})$ denotes the vector of singular values of matrix $\mathbf{X}$ and $\|\cdot\|_p$ is the $\ell_p$ norm with $p\in[0,1]$. As is known, for $p=1$, the Schatten-$p$ norm reduces to the well-known nuclear norm $\|\mathbf{X}\|_{*}$, which has been proven to be the convex envelope of the rank, \cite{fazel2002}. Schatten-$p$ norms have played a significant role in numerous cases involving the rank minimization problem of (\ref{pb_1:af_rank_minim}) reformulating it as
\begin{align}
\mathrm{min} \|\mathbf{X}\|^p_{\mathcal{S}_p} \;\;\;\ s.t \;\;\;\;\ \mathbf{\mathcal{A}}(\mathbf{X})= \mathbf{b} \label{pb_1:schatten_rank_minim}.
\end{align} 

Nowadays, Schatten-$p$ norm based minimization has been seen via a more intriguing perspective i.e. using an iterative reweighting approach. In this vein, inspired by iteratively reweighted least squares (LS) used in place of $\ell_1$ norm minimization for imposing sparsity,\cite{Daubechies}, in \cite{mohan2012iterative,fornasier2011low} the authors propose to minimize a reweighting Frobenious norm, i.e., $\|\mathbf{X}\mathbf{W}^{\frac{1}{2}}\|^2_F$. The equivalence of the Schatten-$p$ norm and the ones minimized in \cite{mohan2012iterative,fornasier2011low}, is mathematically expressed as follows, 
\begin{align}
 \|\mathbf{X}\|^p_{\mathcal{S}_p} &  = \mathrm{tr}\{\left(\mathbf{X}^T\mathbf{X}\right)^{\frac{p}{2}}\}  = \mathrm{tr}\{\left(\mathbf{X}^T\mathbf{X}\right)\left(\mathbf{X}^T\mathbf{X}\right)^{\frac{p-2}{2}}\} \nonumber \\
 & = \mathrm{tr}\{\left(\mathbf{X}^T\mathbf{X}\right)\mathbf{W}\} = \|\mathbf{X}\mathbf{W}^{\frac{1}{2}}\|^2_F,\label{schafro}
\end{align}
where $\mathbf{W}$ is the symmetric weight matrix $\left(\mathbf{X}^T\mathbf{X}\right)^{\frac{p-2}{2}}$. This iterative reweighting scheme has been shown to offer significant merits in terms of the computational complexity of the derived algorithms, the estimation performance as well as the rate of convergence.

Recently, low-rank matrix estimation has been effectively tackled using a {\it matrix factorization} approach. The crux of the relevant methods is that a low-rank matrix can be well represented by a matrix product i.e., $\mathbf{X}=\mathbf{U}\mathbf{V}^T$ with the inner dimension $r$ of the involved matrices quite smaller than the outer dimensions i.e., $r \ll \mathrm{min}(m,n)$. Needless to say that those ideas offer significant advantages when it comes to the processing of large scale and high-dimensional datasets (where both $m$ and $n$ are huge) by reducing the size of the involved variables, thus decreasing both the storage space required  from $\mathcal{O}(mn)$ to $\mathcal{O}\left((m+n)r\right)$ as well as the computational complexity of the algorithms used. However, a downside of this approach is that an additional variable is brought up i.e., the inner dimension $r$ of the factorization. The task of finding the actual $r$ (which coincides with the rank of matrix $\mathbf{X}$) is relevant to the rank minimization problem and is referred in the literature also as dimensionality reduction, model order selection, etc. 

The latter has given rise to methods that select $r$ based on minimization of various criteria such as the Akaike information criterion (AIC), the Bayesian information criterion (BIC), the minimum distance length (MDL), \cite{squires2017rank}, etc. However, these methods can be computationally expensive especially in large scale datasets, since they require multiple runs of the algorithms. Modern approaches termed low-rank matrix factorization (LRMF) techniques, \cite{haeffele2014structured}, hinge on the following philosophy: a) overstate the rank $r$ of the product with $d\geq r$   and then b)  impose low-rankness thereof by utilizing appropriate norms. This rationale has given rise to LRMF techniques that solve the following,
\begin{align}
\mathrm{min}\left[ \mathrm{rank}(\mathbf{U}\mathbf{V}^T) \right] \;\;\ s.t \;\;\ \mathbf{\mathcal{A}}(\mathbf{UV}^T)= \mathbf{b}. \label{LRMF}
\end{align}
Problem (\ref{LRMF}) has been addressed by different ways in literature. Among other approaches, the tight upper-bound of the nuclear norm defined as 
\begin{align}
\|\mathbf{U}\mathbf{V}^T\|_{\ast} & = \underset{\mathbf{U}\in \mathcal{R}^{m\times d},\mathbf{V}\in\mathcal{R}^{n\times d} }{\mathrm{min}} \|\mathbf{U}\|_F\|\mathbf{V}\|_F \nonumber \\
& = \underset{\mathbf{U}\in \mathcal{R}^{m\times d},\mathbf{V}\in\mathcal{R}^{n\times d} }{\mathrm{min}} \frac{1}{2}\left(\|\mathbf{U}\|^2_F + \|\mathbf{V}\|^2_F \right) \label{upper_bound_nuclear}
\end{align}
is the most popular, \cite{srebro2005rank}. In fact, minimization of (\ref{upper_bound_nuclear}) favors low-rankness on $\mathbf{U}$ and $\mathbf{V}$ by inducing smoothness on these matrices. In \cite{shang,shang2016unified}, the authors derive the tight upper-bounds for all Schatten-$p$ norms with $p\in[0,1]$, (Theorem 1, \cite{shang2016unified}) i.e.,
\begin{align}
\|\mathbf{UV}^T\|_{\mathcal{S}_p}^p& = \underset{\mathbf{U}\in\mathcal{R}^{m\times d},\mathbf{V}\in\mathcal{R}^{n\times d}}{\mathrm{min}} \|\mathbf{U}\|_{\mathcal{S}_{2p}} \|\mathbf{V}\|_{\mathcal{S}_{2p}} \nonumber \\
& = \underset{\mathbf{U}\in\mathcal{R}^{m\times d},\mathbf{V}\in\mathcal{R}^{n\times d}}{\mathrm{min}}\frac{1}{2}\left( \|\mathbf{U}\|_{\mathcal{S}_{2p}}^{2p} + \|\mathbf{V}\|_{\mathcal{S}_{2p}}^{2p} \right).
\label{schatten_p_bounds}
\end{align}
Common denominator of the afore-mentioned low-rank matrix factorization approaches is their direct connection with the low-rank imposing Schatten-$p$ norms, since they provide tight upper-bounds thereof. 

In this work we aspire to apply ideas stemming from {\it iterative reweighting methods for low-rank matrix recovery}, to this challenging low-rank matrix factorization scenario. Therefore, generalizing the above-described low-rank promoting norm upper bounds, we propose to minimize the sum of reweighted (as in (\ref{schafro})) Frobenious norms of the individual factors $\mathbf{U}$ and $\mathbf{V}$. Hence, the newly introduced low-rank inducing function is defined as follows,
\begin{align}
 h(\mathbf{U},\mathbf{V}) = \frac{1}{2}\left(\|\mathbf{U}\mathbf{W}_{\mathbf{U}}^{\frac{1}{2}}\|^2_F + \|\mathbf{V}\mathbf{W}_{\mathbf{V}}^{\frac{1}{2}}\|_F^2\right)
 \label{proposed_lr_term}
\end{align}
where the weight matrices $\mathbf{W}_{\mathbf{U}}$ and $\mathbf{W}_{\mathbf{V}}$ are appropriately selected. In the sequel, we adhere to a special instance of (\ref{proposed_lr_term}) which arises by setting $\mathbf{W}_{\mathbf{U}} = \mathbf{W}_{\mathbf{V}}=\mathbf{W}$ with
\begin{align}
 \mathbf{W} & = \mathrm{diag}\Big(\left(\|\boldsymbol{\mathit{u}}_1\|^2_2 + \|\boldsymbol{\mathit{v}}_1\|^2_2\right)^{p-1},\left(\|\boldsymbol{\mathit{u}}_2\|^2_2 + \|\boldsymbol{\mathit{v}}_2\|^2_2\right)^{p-1}, \nonumber \\
 & \dots,\left(\|\boldsymbol{\mathit{u}}_d\|^2_2 + \|\boldsymbol{\mathit{v}}_d\|^2_2\right)^{p-1}   \Big),
\label{weight_matrix}
\end{align}
where $\mathit{\boldsymbol{u}}_i$ and $\mathit{\boldsymbol{v}}_i$ are the $i$th columns of $\mathbf{U}$ and $\mathbf{V}$, respectively\footnote{If $\mathbf{U},\mathbf{V}$ had orthogonal columns, $\mathbf{W}$ in (\ref{weight_matrix})  would be equal to $(\mathbf{U}^T\mathbf{U}+\mathbf{V}^T\mathbf{V})^{p-1}$.}. It can be easily observed that by selecting a common $\mathbf{W}$ for $\mathbf{U}$ and $\mathbf{V}$ as defined in (\ref{weight_matrix}), matrices $\mathbf{U}$ and $\mathbf{V}$ are implicitly coupled w.r.t. their columns. By setting now $p=\frac{1}{2}$ and substituting (\ref{weight_matrix}) in (\ref{proposed_lr_term}) yields
\begin{align}
h(\mathbf{U},\mathbf{V}) = \frac{1}{2}\sum^d_{i=1} \sqrt{\|\mathit{\boldsymbol{u}}_i\|^2_2 + \|\mathit{\boldsymbol{v}}_i\|^2_2}. \label{proposed_lrt}
\end{align}
Surprisingly, the resulting expression coincides with the (scaled by 1/2) group sparsity inducing $\ell_1/\ell_2$ norm of the concatenated matrix $[\begin{smallmatrix} \mathbf{U} \\ \mathbf{V} \end{smallmatrix} ]$ .
Intuitively, the low-rank inducing properties of the proposed in (\ref{proposed_lrt}) joint column sparsity promoting term can be easily explained as follows. Let us consider the rank one decomposition of the matrix product $\mathbf{U}\mathbf{V}^T$,
\begin{align}
\mathbf{U}\mathbf{V}^T = \sum^d_{i=1} \mathit{\boldsymbol{u}}_i\mathit{\boldsymbol{v}}_i^T. \label{rank_one_decomp}
\end{align} 
Clearly, due to the subadditivity property of the rank, eliminating rank one terms of the summation on the right side of (\ref{rank_one_decomp}) results to a relevant decrease of the rank of the product $\mathbf{U}\mathbf{V}^T$. Hence capitalizing on (\ref{proposed_lrt}), we are led to LRMF optimization problems having the form,
\small
\begin{align}
\underset{\mathbf{U}\in\mathcal{R}^{m\times d},\mathbf{V}\in\mathcal{R}^{n\times d}}{\mathrm{min}} \sum^d_{i=1} \sqrt{\|\mathit{\boldsymbol{u}}_i\|^2_2 + \|\mathit{\boldsymbol{v}}_i\|^2_2} \;\ s.t \;\ \mathbf{\mathcal{A}}(\mathbf{UV}^T)= \mathbf{b}. \label{proposed_optim_problem}
\end{align}
\normalsize
It should be noted that the idea of imposing jointly column sparsity first appeared in \cite{tan2009automatic}, albeit in a Bayesian framework tailored to the NMF problem. In \cite{tan2013automatic}, the emerging via the maximum a posteriori probability (MAP) approach optimization problem boils down to the minimization of the column sparsity promoting concave logarithm function. On the other hand, the proposed approach is related to the convex $\ell_1/\ell_2$ norm. The relevance of the proposed formulation to that of the Bayesian schemes proposed in \cite{tan2013automatic} is further highlighted in the next subsection, which describes an instance of problem (\ref{proposed_optim_problem}), as well as two other relevant problems. 

{\it Remark 1: The generic nature of the proposed low-rank promoting function defined in (\ref{proposed_lr_term}) is justified as it includes the previously mentioned MF-based low-rank promoting terms as special cases. Indeed, according to (\ref{schafro}) and by setting $\mathbf{W}_{\mathbf{U}}= (\mathbf{U}^T\mathbf{U})^{p-1}$ and $\mathbf{W}_{\mathbf{V}}= (\mathbf{V}^T\mathbf{V})^{p-1}$ in (\ref{proposed_lr_term}), we get the upper-bound of the Schatten-$p$ norm given in (\ref{schatten_p_bounds}), while for $p=1$, i.e., $\mathbf{W}_{\mathbf{U}} = \mathbf{W}_{\mathbf{V}} = \mathbf{I}_{d}$, we get the variational form of the nuclear norm defined in (\ref{upper_bound_nuclear}).}
\subsection{Denoising, matrix completion and low-rank non-negative matrix factorization}
{\textbf{Denoising}}.
By assuming that a) the linear operator $\mathcal{A}$ reduces to a diagonal matrix and b) our measurements $\mathbf{Y} \in \mathcal{R}^{m\times n}$ are corrupted by i.i.d. Gaussian noise, we come up with the following optimization problem,
\small
\begin{align}
 \underset{\mathbf{U},\mathbf{V}}{\mathrm{min}} \sum^d_{i=1} \sqrt{\|\mathit{\boldsymbol{u}}_i\|^2_2 + \|\mathit{\boldsymbol{v}}_i\|^2_2} \;\;\;\ s.t \;\;\; \|\mathbf{Y} - \mathbf{UV}^T\|^2_F \leq \epsilon. \label{denoising}
\end{align}
\normalsize
where $\epsilon$ is a small positive constant.
By Lagrange theorem we know that (\ref{denoising}) can be equivalently written in the following form,
\footnotesize
\begin{align}
 \{\hat{\mathbf{U}},\hat{\mathbf{V}}\} = \underset{\mathbf{U},\mathbf{V}}{\mathrm{arg min}}\frac{1}{2}\|\mathbf{Y} - \mathbf{UV}^T\|^2_F + {\lambda} \sum^d_{i=1} \sqrt{\|\mathit{\boldsymbol{u}}_i\|^2_2 + \|\mathit{\boldsymbol{v}}_i\|^2_2} \label{proposed_denoising}
\end{align}
\normalsize
where $\lambda$ denotes the Lagrange multiplier.

{\it Proposition 1: The optimization problem (\ref{proposed_denoising}) is equivalent to the MAP minimization scheme arising by placing a Gaussian likelihood on $\mathbf{Y}$ and common, hierarchically formulated, group sparsity promoting Laplace priors on the columns of $\mathbf{U}$ and $\mathbf{V}$.  }

Proposition 1 can be proved following the same steps as those described in the Appendix of \cite{onlineVB2017}. We should point out that in the MAP based schemes of \cite{tan2013automatic}, the prior of $\mathbf{U}$ and $\mathbf{V}$ is the Student-t distribution. For this reason, the corresponding MAP optimization problems involve the concave logarithm function defined on the norms of the columns of $\mathbf{U}$ and $\mathbf{V}$. Contrary, in our case we come up with the $\ell_1/\ell_2$ norm of the matrix resulting by the concatenation of  $\mathbf{U}$ and $\mathbf{V}$. As it is shown later, the simplicity and convexity of the proposed regularizer facilitates not only the derivation of new optimization algorithms, but also the theoretical analysis of their convergence behavior. 

\noindent{\textbf{Matrix completion}}.
Another popular problem that follows the general model described by (\ref{proposed_optim_problem}) is matrix completion, as it is widely addressed via low-rank minimization. The main premise here lies in recovering missing entries of a matrix $\mathbf{Y}$ assuming high coherence among its elements, which gives rise to a low-rank structured matrix $\mathbf{X}$. The problem is thus set up as,
\begin{align}
 \mathrm{min}\left[ \mathrm{rank}(\mathbf{X})\right]\;\;\;\ s.t. \;\;\;\; \mathcal{P}_{\Omega}(\mathbf{Y}) = \mathcal{P}_{\Omega}(\mathbf{X}),
\end{align}
where $\mathcal{P}_{\Omega}$ denotes the sampling operator on the set $\Omega$ of indexes of matrix $\mathbf{Y}$ where  information is present.
In the matrix factorization setting, the incomplete matrix $\mathbf{Y}$ is approximated by a matrix $\mathbf{X}$ expressed as $\mathbf{X}=\mathbf{U}\mathbf{V}^T$. As mentioned above, the rank $r$ of the reconstructed matrix $\mathbf{X}$ is generally unknown and hence it is overstated with $d\geq r$. This necessitates the penalization of the rank of the product $\mathbf{U}\mathbf{V}^T$, which in our case takes place with the proposed low-rank promoting term giving rise to the optimization problem,
\begin{align}
 \underset{\mathbf{U},\mathbf{V}}{\mathrm{min}} \sum^d_{i=1} \sqrt{\|\mathit{\boldsymbol{u}}_i\|^2_2 + \|\mathit{\boldsymbol{v}}_i\|^2_2} \;\;\;\ s.t \;\;\;\;\ \mathcal{P}_{\Omega}(\mathbf{Y}) = \mathcal{P}_{\Omega}(\mathbf{UV}^T) \label{matrix_completion}.
\end{align}
Considering further the existence of additive i.i.d. Gaussian noise in $\mathbf{Y}$ we get, 
\begin{align}
 \{\hat{\mathbf{U}},\hat{\mathbf{V}}\} = & \underset{\mathbf{U},\mathbf{V}}{\mathrm{arg min}}\frac{1}{2}\|\mathcal{P}_{\Omega}(\mathbf{Y}) - \mathcal{P}_{\Omega}(\mathbf{UV}^T)\|^2_F \nonumber \\
 & + \lambda \sum^d_{i=1} \sqrt{\|\mathit{\boldsymbol{u}}_i\|^2_2 + \|\mathit{\boldsymbol{v}}_i\|^2_2}. \label{proposed_mc}
\end{align}

\noindent{\textbf{Low-rank NMF}}.
Finally, we formulate the relevant low-rank constrained non-negative matrix factorization (NMF) problem. The low-rank NMF differs from the classical NMF in the inclusion of the low-rank constraint on the factors $\mathbf{U}$ and $\mathbf{V}$, accounting thus for the unawareness of the true rank. As is shown in Section \ref{sec:experiments} this is very crucial in a class of applications such as music signal decomposition. The emerging optimization problem is given below,
 \begin{align}
 \{\hat{\mathbf{U}},\hat{\mathbf{V}}\} = \underset{\mathbf{U}\geq \mathbf{0},\mathbf{V}\geq \mathbf{0}}{\mathrm{arg min}}& \frac{1}{2}\|\mathbf{Y} - \mathbf{UV}^T\|^2_F \nonumber	 \\ 
 &+ \lambda \sum^d_{i=1} \sqrt{\|\mathit{\boldsymbol{u}}_i\|^2_2 + \|\mathit{\boldsymbol{v}}_i\|^2_2} \label{proposed_low-rank_nmf}
\end{align}
where $\mathbf{U}\geq \mathbf{0}$  and $\mathbf{V}\geq \mathbf{0}$ stand for elementwise non-negativity of $\mathbf{U}$ and $\mathbf{V}$, respectively.
Problem (\ref{proposed_low-rank_nmf}) deviates from the denoising one of (\ref{proposed_denoising}) in the incorporation of an additional contraint i.e., non-negativity of $\mathbf{U},\mathbf{V}$. In the next section three different algorithms, each one solving one of the problems of denoising, matrix completion and low-rank NMF, are developed and theoretically analyzed.
\section{Minimization algorithms}\label{sec:introduction}
Herein, we present three new efficient block coordinate minimization (BCM) algorithms for denoising, matrix completion and low-rank NMF, respectively. The alternating minimization of the proposed low-rank promoting function defined in (\ref{proposed_lrt}) w.r.t. the 'blocks' $\mathbf{U}$ and $\mathbf{V}$ lies at the heart of those algorithms. 

{\it Remark 2: The proposed low-rank promoting regularizer is a) non-smooth and b) non-separable w.r.t. $\mathbf{U}$ and $\mathbf{V}$.}

Both the above-mentioned properties i.e., non-smoothness and non-separability induce severe difficulties in the optimization task that call for appropriate handling. More specifically, as it has been shown, \cite{tseng2001convergence}, in BCM schemes the respective algorithms might be led to irregular points i.e., coordinate-wise minima that are not necessarily stationary points of the minimized cost function. In light of this  we follow a simple smoothing approach by including a small positive constant $\eta$ in the proposed regularizer, which becomes, 
\begin{align}
\hat{h}(\mathbf{U},\mathbf{V}) = \sum^d_{i=1} \sqrt{\|\mathit{\boldsymbol{u}}_i\|^2_2 + \|\mathit{\boldsymbol{v}}_i\|^2_2 + \eta^2}. \label{proposed_low_rank_term}
\end{align}
This way we alleviate singular points i.e., points where the gradient is not continuous, and the resulting optimization problems become smooth. On the other hand, non-separability poses obstacles in getting closed-form expressions for the optimization variables $\mathbf{U}$ and $\mathbf{V}$. For this reason, each of the associative optimization problems is reformulated using appropriate relaxation schemes. By working in an alternating fashion, each of these schemes results in closed form expressions. Next, the proposed algorithms that solve denoising, matrix completion and non-negative matrix factorization are analytically described.
\subsection{Denoising}
In this section, we present a new algorithm designed for solving the denoising problem given in (\ref{proposed_denoising}). To this end, let us first define the respective cost function as, 
\small
\begin{align}
f(\mathbf{U},\mathbf{V}) = \frac{1}{2}\|\mathbf{Y} - \mathbf{U}\mathbf{V}^T\|_F^2 + \lambda \sum^d_{i=1} \sqrt{\|\mathit{\boldsymbol{u}}_i\|^2_2 + \|\mathit{\boldsymbol{v}}_i\|^2_2 + \eta^2}. \label{cost_function_denoising}
\end{align}
\normalsize

It is obvious that minimizing  (\ref{cost_function_denoising}) alternatingly w.r.t. $\mathbf{U}$ and $\mathbf{V}$ is infeasible, since exact analytical expressions can not be obtained as a result of the non-separable nature of the square root. To this end, at each iteration $k+1$ we solve two distinct subproblems i.e. a) given the latest available update $\mathbf{V}_k$ of $\mathbf{V}$, we minimize an approximate cost function w.r.t. $\mathbf{U}$ to get $\mathbf{U}_{k+1}$ and b) we use $\mathbf{U}_{k+1}$ in order to minimize another approximate cost function w.r.t. the second block variable of our problem i.e., matrix $\mathbf{V}$. Following the block successive upper-bound minimization (BSUM) philosophy, \cite{razaviyayn2013unified,hong2016unified}, we minimize at each iteration local tight upper-bounds of the respective cost functions. That said, $\mathbf{U}$ is updated by minimizing an approximate second order Taylor expansion of  $f(\mathbf{U},\mathbf{V}_k)$ around the point $(\mathbf{U}_k,\mathbf{V}_k)$. Likewise, an approximate second-order Taylor expansion of $f(\mathbf{U}_{k+1},\mathbf{V})$ around $(\mathbf{U}_{k+1},\mathbf{V}_k)$ is utilized for obtaining $\mathbf{V}_{k+1}$. To be more specific $\mathbf{U}_{k+1}$ is computed by  
\begin{align}
 \mathbf{U}_{k+1} = \underset{\mathbf{U}}{\mathrm{argmin}}\;\l(\mathbf{U}|\mathbf{U}_k,\mathbf{V}_k), \label{minUk}
\end{align}
where,
\small
\begin{align}
 l(\mathbf{U}|\mathbf{U}_k,\mathbf{V}_k) &= f(\mathbf{U}_k,\mathbf{V}_k) + \mathrm{tr}\{(\mathbf{U}-\mathbf{U}_k)^T\nabla_{\mathbf{U}}f(\mathbf{U}_k,\mathbf{V}_k)\} + \nonumber \\
 &\frac{1}{2}\mathrm{vec}(\mathbf{U}-\mathbf{U}_k)^T\bar{\mathbf{H}}_{\mathbf{U}_k}\mathrm{vec}(\mathbf{U}-\mathbf{U}_k)
 \label{eq:upper_bound_l}
\end{align}
\normalsize
and $\mathrm{vec}(\cdot)$ denotes the row vectorization operator. In (\ref{eq:upper_bound_l}), the true Hessian $\mathbf{H}_{\mathbf{U}_k}$ of $f(\mathbf{U},\mathbf{V}_k)$ at $\mathbf{U}_k$ has been approximated by the $md \times md$ positive-definite block diagonal matrix $\bar{\mathbf{H}}_{\mathbf{U}_k}$, which is expressed as
\begin{equation}
 \bar{\mathbf{H}}_{\mathbf{U}_k} =  \left[ \begin{array}{c c c c} 
            \tilde{\mathbf{H}}_{\mathbf{U}_k} & \mathbf{0} &  \dots & \mathbf{0} \\                       
              \mathbf{0} &   \tilde{\mathbf{H}}_{\mathbf{U}_k} & \ddots & \vdots \\      
              \vdots    & \ddots & \ddots &  \mathbf{0} \\
              \mathbf{0} & \dots &  \mathbf{0} & \tilde{\mathbf{H}}_{\mathbf{U}_k}
              \end{array} \right].
              \label{hessian_bd}
\end{equation} 
In the case of denoising (for reasons that will be explained later) the $d\times d$ diagonal block $\tilde{\mathbf{H}}_{\mathbf{U}_k}$ is defined as
\begin{align}
\tilde{\mathbf{H}}_{\mathbf{U}_k} = \mathbf{V}^T_k\mathbf{V}_k + \lambda\mathbf{D}_{(\mathbf{U}_k,\mathbf{V}_k)} \label{huk}
\end{align}
with 
\begin{align}
 \mathbf{D}_{(\mathbf{U},\mathbf{V})} = \mathrm{diag}\Big(\frac{1}{\sqrt{\|\boldsymbol{\mathit{u}}_1\|^2_2 + \|\boldsymbol{\mathit{v}}_1\|^2_2 + \eta^2}}, \nonumber \\
 \frac{1}{\sqrt{\|\boldsymbol{\mathit{u}}_2\|^2_2 + \|\boldsymbol{\mathit{v}}_2\|^2_2 + \eta^2}},\dots,\frac{1}{\sqrt{\|\boldsymbol{\mathit{u}}_d\|^2_2 + \|\boldsymbol{\mathit{v}}_d\|^2_2 + \eta^2}}\Big).
 \label{definition_D}
\end{align}
As it is shown in the next section, due to the form of $\bar{\mathbf{H}}_{\mathbf{U}_k}$ in (\ref{hessian_bd}) and (\ref{huk}) and its relation to the exact Hessian $\mathbf{H}_{\mathbf{U}_k}$ of $f(\mathbf{U},\mathbf{V}_k)$ at $\mathbf{U}_k$, $l(\mathbf{U}|\mathbf{U}_k,\mathbf{V}_k)$ bounds $f(\mathbf{U},\mathbf{V}_k)$ from above and hence  the conditions set by the BSUM framework are satisfied. Actually, the approximation of the exact Hessian by using (\ref{hessian_bd}) leads to a closed-from expression for updating $\mathbf{U}$ and a dramatic decrease of the required computational complexity, as it will be further explained below. 

Following a similar path as above we come up with appropriate upper-bound functions for updating $\mathbf{V}$ i.e,
\begin{align}
 \mathbf{V}_{k+1} = \underset{\mathbf{V}}{\mathrm{argmin}}\;\ g(\mathbf{V}|\mathbf{U}_{k+1},\mathbf{V}_k) \label{minVk}
\end{align}
with
\footnotesize
\begin{align}
 g(\mathbf{V}|\mathbf{U}_{k+1},\mathbf{V}_k) &= f(\mathbf{U}_{k+1},\mathbf{V}_k) + \mathrm{tr}\{(\mathbf{V}-\mathbf{V}_k)^T\nabla_{\mathbf{V}}f(\mathbf{U}_{k+1},\mathbf{V}_k)\} + \nonumber \\
 &\frac{1}{2}\mathrm{vec}(\mathbf{V}-\mathbf{V}_k)^T\bar{\mathbf{H}}_{\mathbf{V}_k}\mathrm{vec}(\mathbf{V}-\mathbf{V}_k)
 \label{eq:upper_bound_g}
\end{align}
\normalsize
and  $\bar{\mathbf{H}}_{\mathbf{V}_k}$ being a block diagonal $md \times md$ matrix (similar to $\bar{\mathbf{H}}_{\mathbf{U}_k}$) whose $d\times d$ diagonal blocks $\tilde{\mathbf{H}}_{\mathbf{V}_k}$ are defined as
\begin{align}
\tilde{\mathbf{H}}_{\mathbf{V}_k} = \mathbf{U}^T_{k+1}\mathbf{U}_{k+1} + \lambda\mathbf{D}_{(\mathbf{U}_{k+1},\mathbf{V}_{k})}. \label{hvk}
\end{align}
By solving (\ref{minUk}) and (\ref{minVk}) we obtain analytical expressions for $\mathbf{U}_{k+1}$ and $\mathbf{V}_{k+1}$ that constitute the main steps of the proposed denoising algorithm given in Algorithm 1. 

{\it Remark 3: Interestingly, the update formulas for $\mathbf{U}$ and $\mathbf{V}$ derived before could have been derived from  iteratively reweighted least squares (IRLS) minimization schemes \cite{beck2015convergence}.  Indeed, the IRLS algorithm solves (\ref{minUk}) with $l(\mathbf{U}|\mathbf{U}_k,\mathbf{V}_k)$ defined as, 
\small
\begin{align}
l(\mathbf{U}|\mathbf{U}_k,\mathbf{V}_k) = \frac{1}{2} \|\mathbf{Y} - \mathbf{UV}_k^T\|^2_F
+ \frac{\lambda}{2} \sum^d_{i}\frac{\|\mathit{\boldsymbol{u}}_i\|^2_2 + \|\mathit{\boldsymbol{v}}^k_i\|^2_2 + \eta^2}{\sqrt{\|\mathit{\boldsymbol{u}}^k_i\|^2_2 + \|\mathit{\boldsymbol{v}}^k_i\|^2_2 + \eta^2}} \nonumber
\end{align}
\normalsize
and (\ref{minVk})  with a similar definition for  $g(\mathbf{V}|\mathbf{U}_{k+1},\mathbf{V}_k)$. It can be shown that solving these two new optimization problems, we get the same exact closed-form expressions for $\mathbf{U}_{k+1}$ and $\mathbf{V}_{k+1}$ as previously. 
} 

{\it Remark 4: For $\lambda>0$, approximation matrices $\bar{\mathbf{H}}_{\mathbf{U}_k}$ and $\bar{\mathbf{H}}_{\mathbf{V}_k}$ are always positive definite and hence invertible. In other words, both $l(\mathbf{U}|\mathbf{U}_k,\mathbf{V}_k)$ and $g(\mathbf{V}|\mathbf{U}_{k+1},\mathbf{V}_k)$ are strictly convex and hence have unique minimizers. In addition, since approximations of the exact Hessians are used in the two block problems, we end up with quasi-Newton type update formulas for $\mathbf{U}$ and $\mathbf{V}$.}
\begin{table}
\centering
\title{Algorithm 1: Alternating iteratively reweighted least squares (AIRLS) denoising algorithm}
 \begin{tabular}{|l|}
 \hline \\
  Algorithm 1 :Alternating iteratively reweighted least \\
  squares (AIRLS) denoising 
  algorithm\\ \hline 
  Input: $\mathbf{Y},\lambda>0$ \\
  Initialize: $k=0, \mathbf{V}_0,\mathbf{U}_0, \mathbf{D}_{(\mathbf{U}_0,\mathbf{V}_0)}$    \\
  \bf{repeat}\\
    \hspace{0.5cm} $\mathbf{U}_{k+1} = \mathbf{Y}^T\mathbf{V}_{k}\left(\mathbf{V}^T_{{k}}\mathbf{V}_{k} + \lambda \mathbf{D}_{(\mathbf{U}_{k},\mathbf{V}_{k})}\right)^{-1}$ \\
    \hspace{0.5cm} $\mathbf{V}_{k+1} = \mathbf{Y}\mathbf{U}^T_{k+1}\left(\mathbf{U}^T_{k+1}\mathbf{U}_{k+1} + \lambda \mathbf{D}_{(\mathbf{U}_{k+1},\mathbf{V}_{k})}\right)^{-1}   $\\
      \hspace{0.5cm}$k=k+1$\\
    \bf{until} {\it convergence} \\
    Output: $\hat{\mathbf{U}} = \mathbf{U}_{k+1}, \hat{\mathbf{V}} = \mathbf{V}_{k+1}$     \\
    \hline
 \end{tabular}
\end{table}
\vspace{-0.2cm}
\normalsize
\subsection{Matrix completion}
Next the matrix completion problem, under the matrix factorization setting stated in (\ref{proposed_mc}), is addressed. As mentioned earlier, matrix factorization offers scalability making the derived algorithms amenable to processing big and high dimensional data. It should be emphasized that in the proposed formulation of the problem (\ref{proposed_mc}), the impediments arising by the low-rank promoting term (Remark 2) are now complemented by the difficulty to get computationally efficient matrix-wise updates for $\mathbf{U}$ and $\mathbf{V}$, due to the presence of the sampling operator $\mathcal{P}_{\Omega}$ in the data fitting term. That said, the cost function is now modified as
\small
\begin{align}
 f(\mathbf{U},\mathbf{V}) = \frac{1}{2}\|\mathcal{P}_{\Omega}\left(\mathbf{Y} - \mathbf{U}\mathbf{V}^T\right)\|_F^2 
 + \lambda \sum^d_{i=1} \sqrt{\|\mathit{\boldsymbol{u}}_i\|^2_2 + \|\mathit{\boldsymbol{v}}_i\|^2_2 + \eta^2}. \label{cost_function_matrix_completion}
\end{align}
\normalsize

As in the denoising problem, we utilize quadratic upper-bound functions based on approximate second-order Taylor expansions. Again, at each iteration, $\mathbf{U}$ and $\mathbf{V}$ are alternatingly updated by minimizing $l(\mathbf{U}|\mathbf{U}_k,\mathbf{V}_k)$ and $g(\mathbf{V}|\mathbf{U}_{k+1},\mathbf{V}_k)$ defined in (\ref{eq:upper_bound_l}) and (\ref{eq:upper_bound_g}), with $\bar{\mathbf{H}}_{\mathbf{U}_k}$ and $\bar{\mathbf{H}}_{\mathbf{V}_k}$ as given before, but $f(\mathbf{U},\mathbf{V})$ is now defined as in (\ref{cost_function_matrix_completion}). The resulting update formulas are shown in Algorithm 2, where the new AIRLS matrix completion algorithm is presented. 

{\it Remark 5: The gain of using matrices $\bar{\mathbf{H}}_{\mathbf{U}_k}$ and $\bar{\mathbf{H}}_{\mathbf{V}_k}$ in the approximation of the exact Hessians of $f(\mathbf{U},\mathbf{V})$ (given either by (\ref{cost_function_denoising}) or (\ref{cost_function_matrix_completion})) w.r.t. $\mathbf{U}$ and $\mathbf{V}$ is twofold. Not only we remain in the BSUM framework, which offers favorable theoretical properties, but also we are able to update $\mathbf{U}$ and $\mathbf{V}$ at a very low computational cost. As it can be noticed in Algorithms 1 and 2, the inversions of $\bar{\mathbf{H}}_{\mathbf{U}_k}$ and $\bar{\mathbf{H}}_{\mathbf{V}_k}$ involved in the updates of $\mathbf{U}$ and $\mathbf{V}$ reduce to the inversion of the $d\times d$ matrices $\tilde{\mathbf{H}}_{\mathbf{U}_k}$ and $\tilde{\mathbf{H}}_{\mathbf{V}_k}$ thus inducing complexity in the order of $\mathcal{O}(d^3)$. Contrary, utilization of the exact Hessians  w.r.t.  $\mathbf{U}$ and $\mathbf{V}$ would have given rise to inversions with much higher computational complexity i.e., $\mathcal{O}(\mathrm{max}(m,n)\times d^3)$.}
\begin{table}
\centering
\title{Algorithm 1 :Alternating iterative reweighted least squares matrix completion algorithm}
 \begin{tabular}{|l|}
 \hline \\
  Algorithm 2: AIRLS matrix completion (AIRLS-MC) \\
	algorithm \\ \hline
  Input: $\mathbf{Y},\delta$ \\
  Initialize: $k=0, \mathbf{U}_0,\mathbf{V}_0,\mathbf{D}_{(\mathbf{U}_0,\mathbf{V}_0)}$    \\
  \bf{repeat} \\
    \hspace{0.5cm} $\mathbf{U}_{k+1} = \mathbf{U}_{k} - \Big( \mathcal{P}_{\Omega}\left(\mathbf{U}_{k}\mathbf{V}^T_{{k}} -\mathbf{Y}\right)\mathbf{V}_{k} $ \\
   \hspace{0.5cm} $ + \mathbf{U}_{k}\mathbf{D}_{(\mathbf{U}_{k},\mathbf{V}_{k})}\Big)\left(\mathbf{V}^T_{{k}}\mathbf{V}_{k} + \lambda\mathbf{D}_{(\mathbf{U}_{k},\mathbf{V}_{k})}\right)^{-1}$ \\
    \hspace{0.5cm}  $\mathbf{V}_{k+1} = \mathbf{V}_{k} - \Big(\mathcal{P}_{\Omega}\left(\mathbf{V}_{k}\mathbf{U}_{k+1}^T - \mathbf{Y}^T \right)\mathbf{U}_{k+1}$ \\
   \hspace{0.5cm} $  + \mathbf{V}_{k}\mathbf{D}_{(\mathbf{U}_{k+1},\mathbf{V}_{k})}\Big)\left(\mathbf{U}^T_{k+1}\mathbf{U}_{k+1} + \lambda\mathbf{D}_{(\mathbf{U}_{k+1},\mathbf{V}_{k})}\right)^{-1}$\\
   \hspace{0.5cm} $k=k+1$\\
    \bf{until} {\it convergence} \\
    Output: $\hat{\mathbf{U}} = \mathbf{U}_{k+1}, \hat{\mathbf{V}} = \mathbf{V}_{k+1}$     \\
    \hline
 \end{tabular}
 \vspace{-0.4cm}
\end{table}
\vspace{-0.0cm}
\subsection{Non-negative matrix factorization}
In what follows, we present a projected Newton-type method for efficiently addressing the nonnegative matrix factorization problem. It deserves to notice that we are now dealing with a constrained optimization problem since the solution set of the matrices $\mathbf{U}$ and $\mathbf{V}$ contains only elementwise nonnegative matrices. Following the same path presented above we aim at exploiting the curvature information of the formed cost function. However the constrained nature of the NMF problem induces some subtleties needed to be properly handled.

More specifically, the proposed alternating minimization algorithm shall now update matrices $\mathbf{U}$ and $\mathbf{V}$ so that they a) always belong to the feasibility set and b) guarantee the descent direction of the cost function at each iteration. 
The proposed scheme is along the lines of the NMF algorithm proposed in \cite{gong2012efficient}. Each update of the factors takes place making use of the projected Newton method introduced in \cite{bertsekas1982projected}. Next, the minimization subproblems for updating the factors $\mathbf{U}$ and $\mathbf{V}$ are detailed.

As in the previous algorithms, surrogate quadratic functions of $f(\mathbf{U},\mathbf{V}_k)$ and $f(\mathbf{U}_{k+1},\mathbf{V})$ are required for updating matrices $\mathbf{U}$ and $\mathbf{V}$ with $f(\mathbf{U},\mathbf{V})$ being the same as in eq. (\ref{cost_function_denoising}), but now the entries of $\mathbf{U}$ and $\mathbf{V}$ belong to the set of nonnegative reals. 
Let us now consider the so-called set of {\it active constraints} defined  w.r.t. each row $\mathbf{u}_i$ of $\mathbf{U}$ at iteration $k$ as
\begin{align}
 \mathcal{I}^k_{\mathbf{u}_i} = \{j| 0 \leq {u}^k_{ij} \leq \epsilon^k, [\nabla_{{\mathbf{U}}} f(\mathbf{U}_k,\mathbf{V}_k)]_{ij}>0  \},
\end{align}
where $\epsilon^k = \mathrm{min}(\varepsilon,\|\mathbf{U}_k - \nabla_{\mathbf{U}}f(\mathbf{U}_k,\mathbf{V}_k)\|^2_F)$ (with $\varepsilon$ a small positive constant). A similar set $\mathcal{I}^k_{\mathbf{v}_i}$ is defined based on the rows $\mathbf{v}_i$ of matrix $\mathbf{V}$ i.e.,
\small
\begin{align}
 \mathcal{I}^k_{\mathbf{v}_i} = \{j| 0 \leq {v}^k_{ij} \leq \epsilon^k, [\nabla_{{\mathbf{V}}} f(\mathbf{U}_{k+1},\mathbf{V}_k)]_{ij}>0  \}.
\end{align}\normalsize
As is analytically explained in \cite{gong2012efficient}, these sets contain the coordinates of the row elements of matrices $\mathbf{U}$ and $\mathbf{V}$ that belong to the boundaries of the constrained sets, and at the same time are stationary at iteration $k$. 
To derive a projected Newton NMF algorithm, we replace the exact Hessian of each subproblem, with a positive definite matrix that has been partially diagonalized at each iteration w.r.t. the sets of active constraints defined above. The positive definite matrices utilized in this case, denoted as $\bar{\mathbf{H}}^{\mathcal{I}_{\mathbf{U}}}_{\mathbf{U}}$ and $\bar{\mathbf{H}}^{\mathcal{I}_{\mathbf{V}}}_{\mathbf{V}}$, in analogy  to $\bar{\mathbf{H}}_{\mathbf{U}}$ and $\bar{\mathbf{H}}_{\mathbf{V}}$ used in the cases of denoising and matrix completion, are block diagonal, but consist of $m$ and $n$, respectively, $d\times d$ {\it distinct} diagonal blocks. That is to say, the $i$th diagonal blocks of these matrices at iteration $k$, namely $\tilde{\mathbf{H}}^{\mathcal{I}_{\mathbf{u}_i}^k}_{\mathbf{U}}$ and $\tilde{\mathbf{H}}^{\mathcal{I}_{\mathbf{v}_i}^k}_{\mathbf{V}}$, are partially diagonalized versions of the $d\times d$ matrices $\tilde{\mathbf{H}}_{\mathbf{U}_k}$ and $\tilde{\mathbf{H}}_{\mathbf{V}_k}$ defined in (\ref{huk}) and (\ref{hvk}). More specifically, 
\[
 [\tilde{\mathbf{H}}^{\mathcal{I}^k_{\mathbf{u}_i}}_{\mathbf{U}}]_{pl} = \begin{cases} 0, \text{if} \ p\neq l, \text{and either}\  p\in \mathcal{I}^k_{\mathbf{u}_i} \ \text{or}\ l \in \mathcal{I}^k_{\mathbf{u}_i}\\
 [\tilde{\mathbf{H}}_{\mathbf{U}_k}]_{pl}   \  \text{otherwise}
\end{cases}
\]
and $\tilde{\mathbf{H}}^{\mathcal{I}^k_{\mathbf{v}_i}}_{\mathbf{V}}$ is defined similarly. 

Based on the above, the quadratic surrogate functions $l(\mathbf{U}|\mathbf{U}_k,\mathbf{V}_k)$ and $g(\mathbf{V}|\mathbf{U}_{k+1},\mathbf{V}_k)$ are now expressed as,
\small
\begin{align}
 l(\mathbf{U}|\mathbf{U}_k,\mathbf{V}_k) = f(\mathbf{U}_k,\mathbf{V}_k)+ \mathrm{tr}\{\left(\mathbf{U}-\mathbf{U}_k\right)^T\nabla_{\mathbf{U}} f(\mathbf{U}_k,\mathbf{V}_k)\}   \nonumber \\ + \frac{1}{2\alpha^k_{\mathbf{U}} }\mathrm{vec}\left(\mathbf{U}-\mathbf{U}_k\right)^T\bar{\mathbf{H}}^{{\mathcal{I}}^k_{\mathbf{U}}}_{\mathbf{U}}\mathrm{vec}\left(\mathbf{U}-\mathbf{U}_k\right)
\end{align}
\normalsize
and
\small
\begin{align}
 g(\mathbf{V}|\mathbf{U}_{k+1},\mathbf{V}_k)& = f(\mathbf{U}_{k+1},\mathbf{V}_k) +\nonumber \\
  & \mathrm{tr}\{\left(\mathbf{V}-\mathbf{V}_k\right)^T\nabla_{\mathbf{V}} f(\mathbf{U}_{k+1},\mathbf{V}_k)\} +  \nonumber \\ 
 & \frac{1}{2\alpha^k_{\mathbf{V}} }\mathrm{vec}\left(\mathbf{V}-\mathbf{V}_k\right)^T\bar{\mathbf{H}}^{{\mathcal{I}^k_{\mathbf{V}}}}_{\mathbf{V}}\mathrm{vec}\left(\mathbf{V}-\mathbf{V}_k\right),
\end{align}
\normalsize
where  $\alpha^k_{\mathbf{U}}$ and $\alpha_{\mathbf{V}}^k$ denote step size parameters.
Hence, $\mathbf{U}$ and $\mathbf{V}$ are updated by solving the following constrained minimization problems,
\begin{align}
 \mathbf{U}_{k+1} = \underset{\mathbf{U} \geq \mathbf{0}}{\mathrm{argmin}} \ l(\mathbf{U}|\mathbf{U}_k,\mathbf{V}_k) \\
 \text{and}\;\;\;\ \mathbf{V}_{k+1} = \underset{\mathbf{V}\geq \mathbf{0}}{\mathrm{argmin}} \ g(\mathbf{V}|\mathbf{U}_{k+1},\mathbf{V}_k)
\end{align}
giving rise to feasible updates  in the form
\begin{align}
 \mathrm{vec}(\mathbf{U}_{k+1}(\alpha^k_{\mathbf{U}}))& = [\mathrm{vec}(\mathbf{U}_k) - \nonumber \\
 &\alpha^k_{\mathbf{U}} \left(\bar{\mathbf{H}}^{\mathcal{I}_{\mathbf{U}}^k}_{\mathbf{U}}\right)^{-1}\mathrm{vec}(\nabla_\mathbf{U} f(\mathbf{U}_k,\mathbf{V}_k))]_{+} \\
 \mathrm{vec}(\mathbf{V}_{k+1}(\alpha_{\mathbf{V}}^k)) & = [\mathrm{vec}(\mathbf{V}_k) - \nonumber \\ 
 &\alpha^k_{\mathbf{V}} \left(\bar{\mathbf{H}}^{\mathcal{I}_{\mathbf{V}}^k}_{\mathbf{V}}\right)^{-1}\mathrm{vec}(\nabla_\mathbf{V} f(\mathbf{U}_{k+1},\mathbf{V}_k))]_{+},
\end{align}
where $[x]_+ = \max(x,0)$. The step size parameters $\alpha^k_{\mathbf{U}}$ and $\alpha_{\mathbf{V}}^k$ are calculated based on the Armijo rule on the projection arc, \cite{bertsekas1999nonlinear}, with the goal of achieving sufficient decrease of the initial cost function per iteration. Concretely, $\alpha^k_{\mathbf{U}}$ is set to $\alpha^k_{\mathbf{U}} = \beta_{\mathbf{U}}^{{m}_k}$ with $\beta_{\mathbf{U}}\in(0,1)$ and $m_k$ is the first nonnegative integer such that 
\begin{align}
 &f(\mathbf{U}_k) - f(\mathbf{U}_{k+1}(\alpha^k_{\mathbf{U}}))\geq \nonumber \\
 &\sigma\Bigg\{\alpha^k_{\mathbf{U}}\sum_{\footnotesize i \notin \{\mathcal{I}^k_{\mathbf{u}_1} \cup \mathcal{I}^k_{\mathbf{u}_2} \cup\dots \cup \mathcal{I}^k_{\mathbf{u}_m}\}}\frac{\partial f(\mathbf{U}_k,\mathbf{V}_k) }{\partial \mathrm{vec}(\mathbf{U})_i}\times \nonumber \\
& \left(\left(\bar{\mathbf{H}}^{\mathcal{I}^k_{\mathbf{U}}}_{\mathbf{U}}\right)^{-1}\mathrm{vec}(\nabla_\mathbf{U} f(\mathbf{U}_k,\mathbf{V}_k))\right)_i +\nonumber \\
 & \sum_{i\in \{\mathcal{I}^k_{\mathbf{u}_1} \cup \mathcal{I}^k_{\mathbf{u}_2} \cup\dots \cup \mathcal{I}^k_{\mathbf{u}_m}\} } \frac{\partial f(\mathbf{U}_k,\mathbf{V}_k) }{\partial \mathrm{vec}(\mathbf{U})_i}\times 
  \mathrm{vec}(\mathbf{U}_k - \mathbf{U}_k(\alpha^k_{\mathbf{U}}))_i\Bigg\}. \label{armijo_rule}
\end{align}
where $\sigma$ is a constant scalar. The same process described above for selecting $\alpha^k_{\mathbf{U}}$ and hence updating $\mathbf{U}$ is subsequently adopted for $\alpha^k_{\mathbf{V}}$ and $\mathbf{V}$.  The resulting alternating projected Newton-type algorithm for low-rank NMF is given in Algorithm 3. 

{\it Remark 6: The adopted Armijo-rule on the projection arc provides us guarantees regarding the monotonic decrease of the initial cost function per iteration as detailed in the next section. It should be noted that, contrary to the projected Newton NMF method of \cite{gong2012efficient}, in our case the diagonal matrices adopted are always positive definite and hence invertible offering stability to the derived algorithm. Finally, since the approximate Hesssian matrices used are partially diagonal, efficient implementations can be followed for reducing the computational cost.}
\begin{table}
\centering
\title{Algorithm 3: Low-rank nonnegative matrix factorization\\
algorithm }
 \begin{tabular}{|l|}
 \hline \\
  Algorithm 3: AIRLS nonnegative matrix factorizarion \\
	(AIRLS-NMF) algorithm\\ \hline
  Input: $\mathbf{Y},\lambda,\beta_{\mathbf{U}},\beta_{\mathbf{V}},\sigma,\epsilon=10^{-6}$ \\
  Initialize: $k=0, \mathbf{U}^0,\mathbf{V}^0,\mathbf{D}_{(\mathbf{U}_0,\mathbf{V}_0)}$    \\
  \bf{repeat} \\
    \hspace{0.5cm} Estimate the set of active constraints $\mathcal{I}^k_{\mathbf{U}}$\\
    \hspace{0.5cm} $m_k=0$ \\
    \hspace{0.5cm} \textbf{while} \text{eq. (\ref{armijo_rule})} \textbf{do} \\
    \hspace{1cm}  $m_k	=m_k+1$, $\alpha^k_{\mathbf{U}} = \beta_{\mathbf{U}}^{m_k}$\\
    \hspace{0.5cm} \bf{end} \\
    \hspace{0.5cm} $\mathrm{vec}(\mathbf{U}_{k+1}) = [\mathrm{vec}({\mathbf{U}}_k) - $\\
    \hspace{2cm}$\alpha^k_{\mathbf{U}} \left(\bar{\mathbf{H}}^{\mathcal{I}^k_{\mathbf{U}}}_{\mathbf{U}}\right)^{-1}\mathrm{vec}(\nabla_\mathbf{U} f(\mathbf{U}_k,\mathbf{V}_k))]_{+}$\\
   \hspace{0.5cm} Estimate the set of active constraints $\mathcal{I}^k_{\mathbf{V}}$\\
    \hspace{0.5cm} $m_k=0$ \\
    \hspace{0.5cm} \textbf{while} \text{eq. (\ref{armijo_rule})} \textbf{do} \\
    \hspace{1cm}  $m_k=m_k+1$, $\alpha^k_{\mathbf{V}} = \beta_{\mathbf{V}}^{m_k}$\\
    \hspace{0.5cm} \bf{end} \\
    \hspace{0.5cm} $\mathrm{vec}(\mathbf{V}_{k+1}) = [\mathrm{vec}(\hat{\mathbf{V}}_k) -$\\
     \hspace{2cm}$\alpha^k_{\mathbf{V}} \left(\bar{\mathbf{H}}^{\mathcal{I}^k_{\mathbf{V}}}_{\mathbf{V}}\right)^{-1}\mathrm{vec}(\nabla_\mathbf{V} f(\mathbf{U}_{k+1},\mathbf{V}_k))]_{+}$\\
     \hspace{0.5cm} $k=k+1$\\
    \bf{until} {\it convergence} \\
    Output: $\hat{\mathbf{U}} = \mathbf{U}_{k+1}, \hat{\mathbf{V}} = \mathbf{V}_{k+1}$     \\
    \hline
 \end{tabular}
 \vspace{-0.2cm}
\end{table}

\vspace{0.1cm}
{\it Remark 7: The proposed AIRLS, AIRLS-MC and AIRLS-NMF algorithms  annihilate jointly columns of the matrices $\mathbf{U}$ and $\mathbf{V}$, as a result of the column sparsity imposing nature of the introduced low-rank promoting term. This key feature of the proposed algorithms let us incorporate a mechanism which prunes the columns that are zeroed as the algorithms evolve. By doing so, the per iteration computational complexity of the algorithms is gradually reduced, and this reduction may become significant, as is also highlighted in the experimental section.}
\vspace{-0.3cm}
\section{Convergence analysis}
In this part of the paper we analyze the convergence behavior of the three algorithms presented in the previous section. Towards this, we first prove the following Lemma.

{\it Lemma 1: The surrogate functions $l(\mathbf{U}|\mathbf{U}_k,\mathbf{V}_k)$ and $g(\mathbf{V}|\mathbf{U}_{k+1},\mathbf{V}_k)$ minimized at each iteration of Algorithms 1 and 2  are tight upper-bounds of the corresponding $f(\mathbf{U},\mathbf{V}_k)$ and $f(\mathbf{U}_{k+1},\mathbf{V})$ with $f(\mathbf{U},\mathbf{V})$ defined in eqs. (\ref{cost_function_denoising}) and (\ref{cost_function_matrix_completion}) for the two algorithms, respectively.}\\
{\it Proof}: See Appendix.

In non-negative matrix factorization, the proposed alternating projected Newton algorithm relies on the approximate Hessians $\bar{\mathbf{H}}^{\mathcal{I}^k_{\mathbf{U}}}_{\mathbf{U}}$ and   $\bar{\mathbf{H}}^{\mathcal{I}^k_{\mathbf{V}}}_{\mathbf{V}}$ defined in the previous section. The following Lemma provides the conditions that ensure that this approach can also be placed within the upper-bound minimization framework. 

{\it Lemma 2: The surrogate function $l(\mathbf{U}|\mathbf{U}_k,\mathbf{V}_k)$ upper bounds $f(\mathbf{U},\mathbf{V}_k)$, if $a^k_{\mathbf{U}}$  is bounded above by $\frac{\lambda_{min}(\bar{\mathbf{H}}^{\mathcal{I}^k_{\mathbf{U}}}_{\mathbf{U}})}{\lambda_{max}({\mathbf{H}}_{\mathbf{U}_k})}$. Similarly, $g(\mathbf{V}|\mathbf{U}_{k+1},\mathbf{V}_k)\geq f(\mathbf{U}_{k+1},\mathbf{V})$, if  $a^k_{\mathbf{V}} \leq \frac{\lambda_{min}(\bar{\mathbf{H}}^{\mathcal{I}^k_{\mathbf{V}}}_{\mathbf{V}})}{\lambda_{max}({\mathbf{H}}_{\mathbf{V}_k})}$, respectively.}\\[0cm]
{\it Proof}: See Appendix.
\\[0cm]
Having shown that the proposed surrogate cost functions are upper bounds of the actual ones, in Proposition 2 given below the monotonic decrease of the initial cost functions per iteration of the respective algorithms is established. 

{\it Proposition 2: The sequences of $\{\mathbf{U}_k,\mathbf{V}_k\}$ generated by Algorithms 1, 2 and 3 decrease monotonically the respective cost functions i.e.,
\small
\begin{equation}
 f(\mathbf{U}_{k+1},\mathbf{V}_{k+1}) \leq f(\mathbf{U}_{k+1},\mathbf{V}_k) \leq f(\mathbf{U}_k,\mathbf{V}_k).
 \label{eq:prop_1}
\end{equation}}
\normalsize
\\[-0.0cm]
{\it Proof:} See Appendix. 

 {\it Corolarry 1: The monotonically decreasing sequence of $f(\mathbf{U}_k,\mathbf{V}_k)$ converges as $k\rightarrow \infty$ to $f^{\infty}\geq 0$.} \\
 {\it Proof:} It can be easily proved using Proposition 2, since the cost functions are bounded below by 0. 
 \vspace{-0.1cm}
\subsection{Rates of convergence and convergence to stationary points}
Having shown that the updates $(\mathbf{U}_k,\mathbf{V}_k)$ generated by Algorithms 1, 2 and 3 monotinically decrease the corresponding cost functions, we herein derive the rates of convergence of the algorithms to a stationary point. The subsequent analysis is along the lines of the one presented in \cite{hastie2015matrix}. 

Given any $(\mathbf{U}, \mathbf{V})$ we define matrices $\mathbf{U}_{\ast}, \mathbf{V}_{\ast}$ arising by the following minimization problems 
\begin{align}
\mathbf{U}_{\ast} = \underset{\mathbf{U}^{+}}{\mathrm{arg min}} \;\;\ l(\mathbf{U}^{+}|\mathbf{U},\mathbf{V} ) \\
\mathbf{V}_{\ast} = \underset{\mathbf{V}^{+}}{\mathrm{arg min}} \;\;\ g(\mathbf{V}^{+}|\mathbf{U}_{*},\mathbf{V} ).
\end{align}
Let us now denote as $\Delta^a((\mathbf{U},\mathbf{V}),(\mathbf{U}_{\ast},\mathbf{V}_{\ast}))$ and $\Delta^b((\mathbf{U},\mathbf{V}),(\mathbf{U}_{\ast},\mathbf{V}_{\ast}))$ the measures of proximity between $(\mathbf{U},\mathbf{V})$  and $(\mathbf{U}_{\ast},\mathbf{V}_{\ast})$ which are defined as follows, 
\begin{align}
 & \Delta^a((\mathbf{U},\mathbf{V}),(\mathbf{U}_{\ast},\mathbf{V}_{\ast})) = 
 \frac{1}{2}\Big(\|\mathbf{V}\left(\mathbf{U} - \mathbf{U}_{\ast}\right)^T\|^2_F +  \nonumber \\
& \|\mathbf{U}_{\ast}\left(\mathbf{V}-\mathbf{V}_{\ast}\right)^T\|^2_F   \Big) + \frac{\lambda}{2}\Big(\|\mathbf{D}_{(\mathbf{U},\mathbf{V})}^{\frac{1}{2}}\left(\mathbf{U} - \mathbf{U}_{\ast}\right)^T\|^2_F + \nonumber \\
&  \|\mathbf{D}_{(\mathbf{U}_{\ast},\mathbf{V})}^{\frac{1}{2}}\left(\mathbf{V} - \mathbf{V}_{\ast}\right)^T\|^2_F \Big) \\
& \Delta^b((\mathbf{U},\mathbf{V}),(\mathbf{U}_{\ast},\mathbf{V}_{\ast})) = \nonumber \\
&\frac{1}{2}\sum^m_{i=1}\left(\mathbf{u}_{i} - \mathbf{u}_{i,\ast}\right)^T[\mathbf{V}^T\mathbf{V}]_{\mathcal{I}_{\mathbf{u}_i}}\left(\mathbf{u}_{i} - \mathbf{u}_{i,\ast}\right) \nonumber \\
& + \frac{1}{2}\sum^n_{i=1} \left(\mathbf{v}_{i}-\mathbf{v}_{i,\ast}\right)^T[\mathbf{U}^T_{\ast}\mathbf{U}_{\ast}]_{\mathcal{I}_{\mathbf{v}_i}}\left(\mathbf{v}_{i}-\mathbf{v}_{i,\ast}\right) \nonumber \\ 
& + \frac{\lambda}{2}\Big(\|\mathbf{D}_{(\mathbf{U},\mathbf{V})}^{\frac{1}{2}}\left(\mathbf{U} - \mathbf{U}_{\ast}\right)^T\|^2_F + \nonumber \\
&\|\mathbf{D}_{(\mathbf{U}_{\ast},\mathbf{V})}^{\frac{1}{2}}\left(\mathbf{V} - \mathbf{V}_{\ast}\right)^T\|^2_F \Big)  + \mathrm{tr}\{(\mathbf{U}-\mathbf{U}_{\ast})^T\nabla_{\mathbf{U}}f(\mathbf{U},\mathbf{V}) +  \nonumber\\ 
&\mathrm{tr}\{(\mathbf{V}-\mathbf{V}_{\ast})^T\nabla_{\mathbf{V}}f(\mathbf{U}_\ast,\mathbf{V})     \}
\end{align}
where $[\mathbf{V}^T\mathbf{V}]_{\mathcal{I}_{\mathbf{u}_i}}$ and $[\mathbf{U}^T_{\ast}\mathbf{U}_{\ast}]_{\mathcal{I}_{\mathbf{v}_i}}$ are partially diagonalized versions of matrices $\mathbf{V}^T\mathbf{V}$ and $\mathbf{U}^T_{\ast}\mathbf{U}_{\ast}$ according to $\mathcal{I}_{\mathbf{u}_i}$ and $\mathcal{I}_{\mathbf{v}_i}$ respectively. 

{\it Lemma 3: Successive differences in the objective values of cost functions $f(\mathbf{U},\mathbf{V})$ corresponding to Algorithms 1,2 and 3 are bounded below as follows,

 For Algorithms 1 and 2: 
 \footnotesize  
\begin{align}
 f(\mathbf{U}_k,\mathbf{V}_k) - f(\mathbf{U}_{k+1},\mathbf{V}_{k+1}) \geq \Delta^a((\mathbf{U}_k,\mathbf{V}_k),(\mathbf{U}_{k+1},\mathbf{V}_{k+1})) \label{lemma_2_a}
\end{align}\normalsize
 For Algorithm 3: 
 \footnotesize
 \begin{align}
 f(\mathbf{U}_k,\mathbf{V}_k) - f(\mathbf{U}_{k+1},\mathbf{V}_{k+1}) \geq  \Delta^b((\mathbf{U}_k,\mathbf{V}_k),(\mathbf{U}_{k+1},\mathbf{V}_{k+1})). \label{lemma_2_b}
\end{align}
\normalsize
} 
{\it Proof:} See Appendix. 

{\it Lemma 4: $\Delta^a((\mathbf{U},\mathbf{V}),(\mathbf{U}_{\ast},\mathbf{V}_{\ast}))=0$ if and only if $(\mathbf{U},\mathbf{V})$ generated by each of the Algorithms 1 and 2, is a  fixed point of them. Likewise,  $\Delta^b((\mathbf{U},\mathbf{V}),(\mathbf{U}_{\ast},\mathbf{V}_{\ast}))=0$ if and only if ($\mathbf{U},\mathbf{V}$) generated by Algorithm 3 is also a fixed point.}\\
{\it Proof:} See Appendix.

Note that $\Delta^a((\mathbf{U}_k,\mathbf{V}_k),(\mathbf{U}_{k+1},\mathbf{V}_{k+1}))$ and $\Delta^b((\mathbf{U}_k,\mathbf{V}_k),(\mathbf{U}_{k+1},\mathbf{V}_{k+1}))$ are actually used for quantifying the distance between $(\mathbf{U}_k,\mathbf{V}_k)$ and $(\mathbf{U}_{k+1},\mathbf{V}_{k+1})$ generated in successive iterations of the proposed algorithms. Thus, it is obvious that if the algorithms converge these measures will become equal to zero. For ease of notation, we will next denote these quantities as $\delta^a_k$ and $\delta^b_k$ respectively. Before proceeding further, we make the following assumption. \\
{\it Assumption 1: The eigenvalues of both $\mathbf{U}^T_k\mathbf{U}_k$  and $\mathbf{V}^T_k\mathbf{V}_k$ for $k\geq 1$ are uniformly bounded below and above by $l_L$ and $l_U$ respectively, i.e.,
\small
\begin{align}
l_{L}\mathbf{I}_d \preceq \mathbf{U}_k^T\mathbf{U}_k \preceq l_{U}\mathbf{I}_d \;\;\; \text{and} \;\;\; l_{L}\mathbf{I}_d \preceq \mathbf{V}_k^T\mathbf{V}_k \preceq l_{U}\mathbf{I}_d.
\end{align}
\normalsize
}
That said, the main result of this section is summarized in the following proposition.

{\it Proposition 3: The sequences of $\{\mathbf{U}_k,\mathbf{V}_k\}$ generated by Algorithms 1,2, and 3 are bounded and hence have at least a limit point. This implies (by Bolzano-Weistrass theorem) that there exist subsequences that converge to the limit points. Actually, the  limit points correspond to fixed points of the Algorithms 1,2 and 3, which are stationary points of the minimized cost functions. Finally, Algorithms 1,2 and 3 converge sublinearly, with their rates of convergence expressed as, 
\begin{align}
 \text{Algorithms 1,2} \;\;\;\; \underset{1\leq k \leq K}{\mathrm{min}}\delta_k^a \leq \frac{f(\mathbf{U}_1,\mathbf{V}_1) - f^{\infty}}{K}  \\
  \text{Algorithm 3} \;\;\;\; \underset{1\leq k \leq K}{\mathrm{min}} \delta_k^b \leq \frac{f(\mathbf{U}_1,\mathbf{V}_1) - f^{\infty}}{K}.
\end{align}

Proof:} See Appendix.

Using Assumption 1 we can provide more refined information with regard to the rates of convergence, bringing into play the curvature characteristics of the cost functions as well as the regularization parameter $\lambda$. 

{\it Corollary 2:  Under Assumption 1, we can derive the following convergence rate for Algortithms 1,2 and 3:
\begin{align}
 \underset{1\leq k \leq K}{\mathrm{min}} \|\mathbf{U}_{k+1}-\mathbf{U}_k\|^2_F + \|\mathbf{V}_{k+1}-\mathbf{V}_k\|^2_F \leq \nonumber \\
 \frac{4\tau}{2l_{L}\tau + \lambda} \frac{f(\mathbf{U}_1,\mathbf{V}_1) - f^{\infty}}{K},
 \label{convergence_rate}
\end{align}
where $\tau = \underset{1\leq i \leq d}{\mathrm{max}}(\|\boldsymbol{\mathit{u}}_i\|^2_2,\|\boldsymbol{\mathit{v}}_i\|^2_2)$.
}\\
{\it Proof:} It can be easily proved by suitably modifying $\delta^a_k$ and $\delta^b_k$ using the inequalities $l_{L}\|\mathbf{U}_{k}-\mathbf{U}_{k+1}\|^2_F \leq \|\mathbf{V}_k\left(\mathbf{U}_{k}-\mathbf{U}_{k+1}\right)\|^2_F \leq l_{U}\|\mathbf{U}_{k}-\mathbf{U}_{k+1}\|^2_F$ and  $l_{L}\|\mathbf{V}_{k}-\mathbf{V}_{k+1}\|^2_F \leq \|\mathbf{U}_{k+1}\left(\mathbf{V}_{k}-\mathbf{V}_{k+1}\right)\|^2_F \leq l_{U}\|\mathbf{V}_{k}-\mathbf{V}_{k+1}\|^2_F$.
\vspace{-0.1cm}
\section{Experiments}\label{sec:experiments}
Next simulated and real data experiments are provided for illustrating the key features of the proposed AIRLS, AIRLS-MC and AIRLS-NMF algorithms. For comparison purposes, the Maximum-Margin-Matrix Factorization (MMMF)
method of \cite{rennie2005fast} is utilized in the denoising type problems. In matrix completion experiments the softImpute-ALS algorithm, \cite{hastie2015matrix}, is used. 
Finally, the ARD-NMF algorithm, \cite{tan2013automatic} is included in the non-negative matrix factorization type experiments. It should be noted that for the three proposed algorithms {\it a column pruning mechanism is applied}. That is, when a column of the matrix factors  has been (approximately) zeroed, it is removed, thus reducing the column size of the factors (see Remark 7). As a result, the per iteration complexity is being reduced during the execution of the algorithms. \vspace{-0.1cm}
\subsection{Simulated data experiments}
Herein we highlight the benefits of the proposed AIRLS, AIRLS-MC and AIRLS-NMF algorithms on simulated data. To this end, the proposed algorithms are tested on two different 
experimental setups i.e. a) for checking the performance of AIRLS and AIRLS-NMF in the presence of noise and b) for testing the capacity of AIRLS-MC in dealing with different percentages of missing data.
\vspace{-0.1cm}
\begin{table*}
 \centering
 \begin{tabular}{|c |c | c |c| c| c| c| c| c| c| c| c| c|}
 \hline 
	SNR &  \multicolumn{6}{c|}{10}& \multicolumn{6}{c|}{20} \\ \hline
    rank &  \multicolumn{3}{c}{5} & \multicolumn{3}{|c}{10} & \multicolumn{3}{|c}{5}& \multicolumn{3}{|c |}{10}  \\\hline
   Algorithm & \small \# Iter & \small time(s) &\small NRE  & \small \# Iter & \small time(s) &\small NRE  & \small \# Iter & \small time(s) &  \small NRE  &\small  \# Iter &\small  time(s) &\small  NRE  \\ \hline
  \multicolumn{1}{|c|}{ MMMF} & 15  & 0,2774 & 0,1079&  15 & 0,2853 & 0,1152 & 40,31 & 0,7739  &0,0235 & 40,38 & 0,7666 & 0,0294\\ \hline
   \multicolumn{1}{|c|}{AIRLS} & 43,37  & 0,3949  & 0,0448  & 24,37 & 0,2426   & 0,0635  & 15,41  & 0,1571 &0,0142 & 35,68  & 0,3421  & 0,02 \\
   \hline
 \end{tabular}
 \caption{Results obtained by MMMF and AIRLS on the simulated denoising experiment.}
 \vspace{-0.5cm}
 \label{table:results_denoising}
\end{table*}
\begin{table*}
 \centering
 \vspace{-0.1cm}
 \begin{tabular}{| c |c | c |c| c| c| c| c| c| c| c| c| c|}
 \hline 
	SNR &  \multicolumn{4}{c|}{10}& \multicolumn{4}{c|}{20} \\ \hline
    rank &  \multicolumn{2}{c}{5} & \multicolumn{2}{|c}{10} & \multicolumn{2}{|c}{5}& \multicolumn{2}{|c|}{10}  \\\hline
   Algorithm & est. rank &\small NRE  & est. rank & \small NRE  & est. rank &\small NRE & est. rank &\small NRE  \\ \hline
  \multicolumn{1}{|c|}{ ARD-NMF} & 4,36  & 0,0778 & 100 & 0,1023 &4,66  & 0,0825 & 100 & 0,1008 \\ \hline
   \multicolumn{1}{|c|}{AIRLS-NMF} & 5,14  & 0,048  & 10,25  & 0,0706& 6,52 & 0,0181 & 10,23& 0,0291\\
   \hline
 \end{tabular}
 \caption{Results obtained by ARD-NMF and AIRLS on the simulated NMF experiment.}
  \vspace{-0.7cm}
 \label{table:results_NMF}
 \vspace{-0.0cm}
\end{table*}
\vspace{-0.0cm}
\subsubsection{AIRLS and AIRLS-NMF}
In order to validate the performance of AIRLS and AIRLS-NMF in the presence of noise two different experimental settings are used. In both settings, a  matrix $\mathbf{X}_{0}\in \mathbb{R}^{m\times n}$  with $m=500$, $n=500$ and varying rank $r \in \{5,10 \}$ is randomly generated. Concretely, matrix $\mathbf{X}_0$ is produced by the product of two matrices i.e., $\mathbf{U}_0\in \mathbb{R}^{m\times r}$ and $\mathbf{V}_0^T\in \mathbb{R}^{r \times n}$  having either a) zero-mean Gaussian entries of variance 1 or b) uniformly distributed non-negative entries in the range 0 to 1. The latter is used for testing the NMF algorithms. In both cases additive Gaussian i.i.d noise 
of different $\mathrm{SNR}\in \{10,20\}$ corrupts $\mathbf{X}_0$, thus resulting to the data matrix $\mathbf{Y}$, which is then provided as input to the tested algorithms. For the case of a) AIRLS is compared to the MMMF algorithm while in b) the ARD-NMF  algorithm takes part in the respective experiments. Note that for the case of  ARD-NMF of \cite{tan2013automatic}, the beta function of its data fitting term is reduced to the squared Frobenious norm. This way, both AIRLS-NMF and ARD-NMF rely on the same noise assumptions.
As a quantitative metric we utilize the normalized reconstruction error defined as 
$\mathrm{NRE} = \frac{\|\mathbf{X}_0- \hat{\mathbf{U}}\hat{\mathbf{V}}^T\|_F}{\|\mathbf{X}_0\|_F}$.  Since we are interested in the recovery performance of the algorithms, the low-rank promoting parameter $\lambda$ of the algorithms is selected from a set of values \{0.1,1,5,10,50,80,100,200\} via fine tuning in terms of the lowest achieved NRE. Moreover, for AIRLS-NMF we set $\beta_{\mathbf{U}}=\beta_{\mathbf{V}}=10^{-1}$ and $\sigma=10^{-2}$. The algorithms stop when either the relative decrease of the reconstructed data between two successive iterations i.e., $\frac{\|\hat{\mathbf{U}}_{k}\hat{\mathbf{V}}_{k}^T - \hat{\mathbf{U}}_{k+1}\hat{\mathbf{V}}_{k+1}^T\|_F}{\|\hat{\mathbf{U}}_{k}\hat{\mathbf{V}}_{k}^T\|_F}$ becomes less than $10^{-4}$ or 500 iterations are reached. 100 independent runs are performed for each algorithm and the average values of the various quantities (elapsed time, NRE, iterations executed and estimated rank) are provided in Tables \ref{table:results_denoising} and \ref{table:results_NMF}. The initial rank is set to $d=100$.

In Table \ref{table:results_denoising}, the results of AIRLS and MMMF are given. Therein, it is shown that AIRLS offers better estimation performance than MMMF in all experiments. Interestingly, in most cases, this happens in less time than that spent by MMMF, although AIRLS in some instances required more iterations. This favorable characteristic of AIRLS is due to its {\it column pruning capability, which results to a much less average time per iteration}. 
In the case of the NMF problem, it can be observed by Table \ref{table:results_NMF} that AIRLS-NMF achieved lower NRE than that of ARD-NMF for all different choices of noise and rank of the sought matrices. Notably, AIRLS-NMF exhibited robustness in recovering the true rank in both cases examined i.e., $r \in \{5,10\}$, contrary to ARD-NMF which failed to estimate the true rank especially for $r=10$. \vspace{-0.1cm}
\subsubsection{AIRLS-MC}
To evaluate the performance of AIRLS-MC in different scenarios, we classify the experimental settings of this subsection according to the degrees of freedom ratio (FR), \cite{mohan2012iterative}, defined as 
$\mathrm{FR} = r (2n - r)/\mathrm{card}(\Omega)$. Recovery becomes harsher as FR is close to 1, whereas easier problems arise when it takes values close to 0. AIRLS-MC is compared to softImpute-ALS for FR equal to $0.4$ and $0.6$. In both cases a low-rank matrix
$\mathbf{X}_{0}\in \mathbb{R}^{m\times n}$  with $m=1000$, $n=1000$ and rank $r=20$ is generated. The NRE defined above is used as the performance metric. For both algorithms, parameter $\lambda$ is fine tuned as described in the previous experiment and the initial rank is set to 100. Again, the algorithms run for 100 instances of each experiment and the mean values of iterations, NRE and time to converge are given in Table \ref{table:results-MC}. Moreover, the same stopping criteria mentioned previously are utilized.
As is shown in Table \ref{table:results-MC}, AIRLS-MC offers higher accuracy than softImpute-ALS in both experiments. Interestingly, this happens in less time, although for  FR=$0.6$ it requires more iterations to converge. Actually, this happens due to the fact that AIRLS-MC estimates the true rank of the matrix after a few iterations. That is, the column pruning mechanism mentioned above reduces gradually its computational complexity. \vspace{-0.1cm}
\begin{table}
 \centering
 \vspace{-0.cm}
 \resizebox{0.5\textwidth}{!}{
 \begin{tabular}{| c |c | c |c| c| c| c| c| c| c| c| c| c|}
 \hline 
	FR &  \multicolumn{3}{c|}{0.4}& \multicolumn{3}{c|}{0.6} \\ \hline
   Algorithm & \small \# Iter & \small time(s) &\small NRE  & \small \# Iter & \small time(s) &\small NRE    \\ \hline
  \multicolumn{1}{|c|}{softImpute-ALS}& 295 & 218 & 0,1851 & 220 & 228 & 0,64 \\ \hline
   \multicolumn{1}{|c|}{AIRLS-MC} & 207  & 53  & 0,1499  & 731  & 174 & 0,27 \\
   \hline
 \end{tabular}}
 \caption{Results of AIRLS-MC and softImpute-ALS on matrix completion experiment.}
 \label{table:results-MC}
 \vspace{-1cm}
\end{table}
\vspace{-0.1cm}
\subsection{Real data experiments}
In this section we validate the performance of the proposed algorithms on three different real data experiments. First, 
the AIRLS algorithm is tested in denoising a real hyperspectral image (HSI). Second, a collaborative filtering application is used 
for testing the matrix completion algorithms. Finally, a music signal decomposition problem is employed for comparing the performance of NMF algorithms. 
\begin{figure}
\begin{tabular}{c}
\begin{tabular}{c c}
 \includegraphics[width=0.22\textwidth,height=0.22\textwidth]{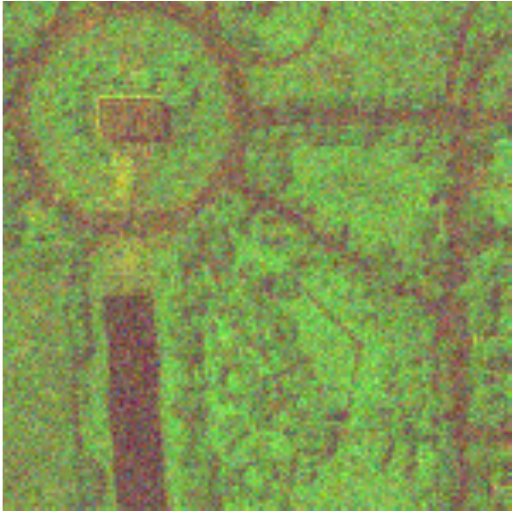} & \includegraphics[width=0.22\textwidth,height=0.22\textwidth]{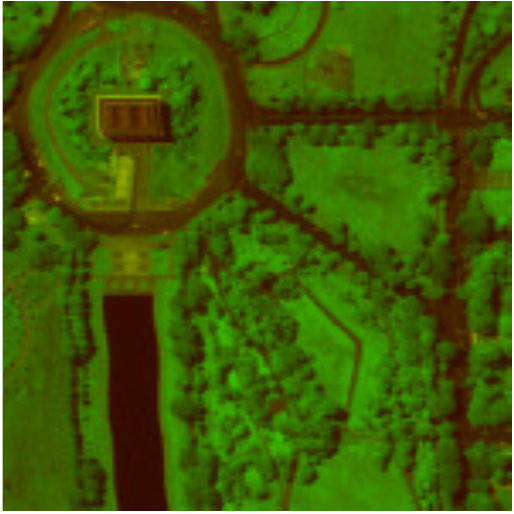}\\
 a)  noisy image&  b) ground truth \\
 \includegraphics[width=0.22\textwidth,height=0.22\textwidth]{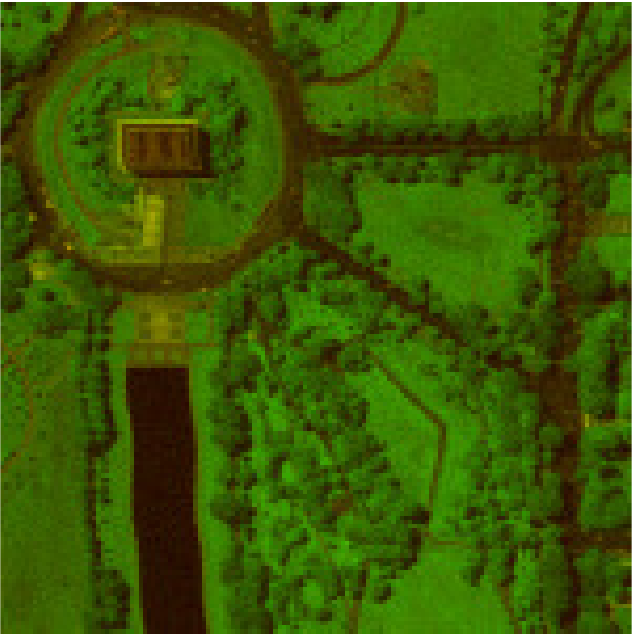} &  \includegraphics[width=0.22\textwidth,height=0.22\textwidth]{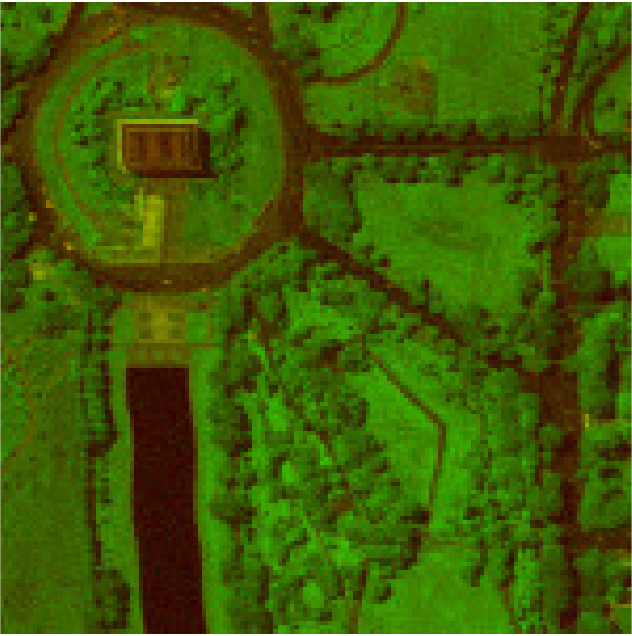}  \\
 c)  MMMF   & d) AIRLS 
 \end{tabular} \\
 \includegraphics[width=0.45\textwidth,height=0.25\textwidth]{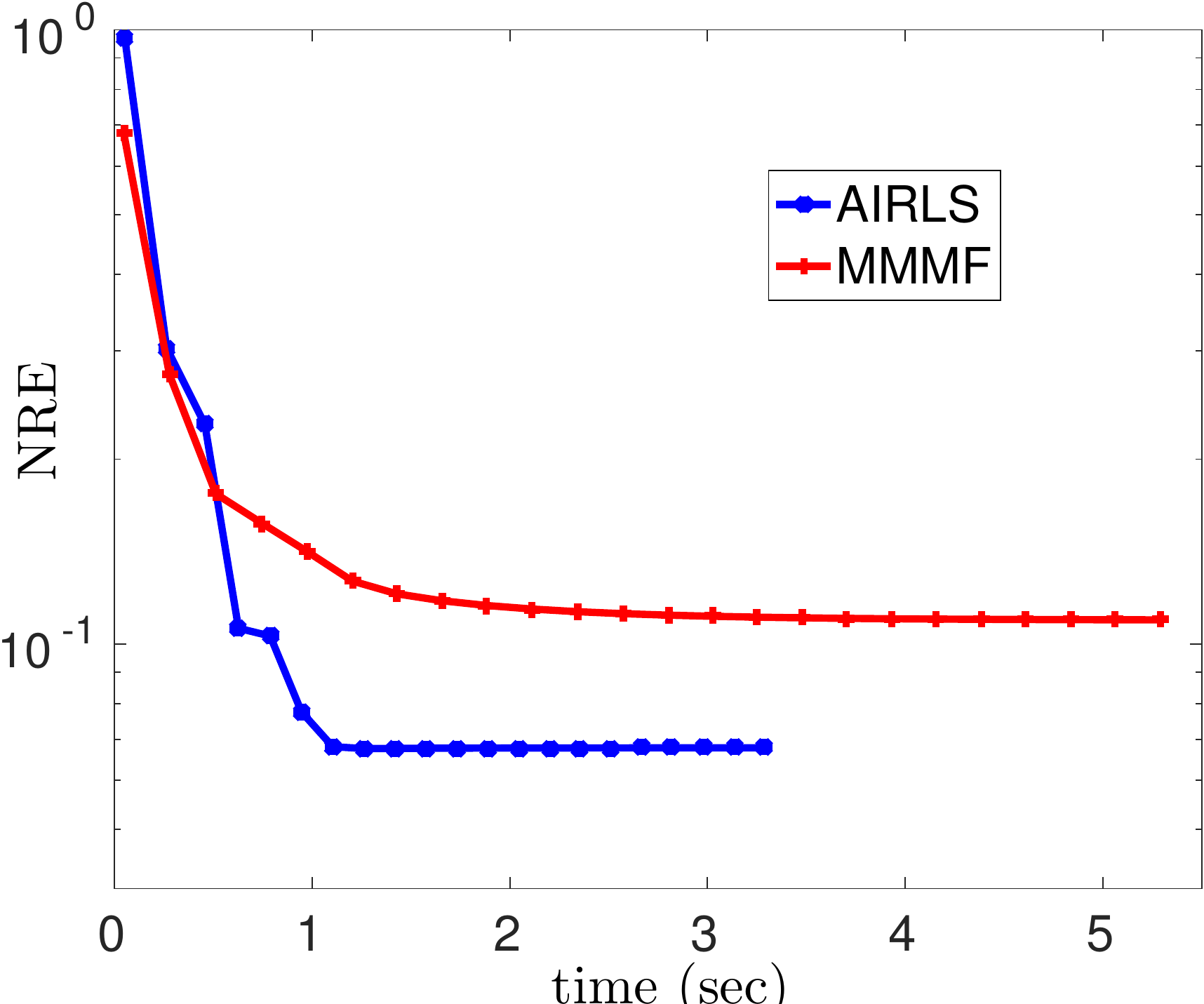} \\
 e) NRE vs time
 \end{tabular}
 \vspace{-0.0cm}
 \caption{Evaluation of AIRLS and MMMF on the Washigton DC AVIRIS dataset.}\vspace{-0.6cm}
 \label{fig:denoising_real_HSI}
\end{figure}
\subsubsection{Hyperspectral Image Denoising}
In this experiment we utilize the Washigton DC Mall AVIRIS HSI captured at $m=210$ contiguous spectral bands in the 0.4 to 2.4 $\mu m$
region of the visible and infrared spectrum. 
The HSI consists of $n= 22500 \ (150\times 150)$ pixels. As is widely known, \cite{giampouras2016simultaneously}, hyperspectral data are highly coherent both in the  spectral and the spatial domains. Therefore, by organizing the tested image in a matrix, whereby each column corresponds to the  spectral bands and each row to the pixels, it turns out that this matrix
can be well approximated by a low-rank one. This fact motivates us to exploit the low-rank structure of the HSI under study for efficiently denoising a highly corrupted version
thereof by Gaussian i.i.d noise of $\mathrm{SNR}=6dB$. 
 
In Fig. \ref{fig:denoising_real_HSI}, false RGB images of the recovered HSIs by the proposed AIRLS algorithm and MMMF are provided. In both algorithms, the number of columns of the initial factors $\mathbf{U}_0$ and $\mathbf{V}_0$ is overstated to $d=100$ and the algorithms terminate when the relative decrease of the reconstructed HSI between two successive iterations reaches a value less than $10^{-4}$. Moreover, their low-rank promoting parameter $\lambda$ is selected so as to lead to solution matrices
$\hat{\mathbf{U}}$ and $\hat{\mathbf{V}}$ of the same rank $r= 4$. As it can be noticed in Fig. \ref{fig:denoising_real_HSI}, AIRLS reconstructs the HSI in a significantly improved accuracy as compared to MMMF. This can be easily verified both by visually inspecting Figs. \ref{fig:denoising_real_HSI}a-\ref{fig:denoising_real_HSI}d and quantitatively in terms of the estimated NRE (Fig. \ref{fig:denoising_real_HSI}e). Notably, AIRLS converges in less iterations than those required by
MMMF (Fig. \ref{fig:denoising_real_HSI}e), while at the same time less time per iteration is consumed, on average. The latter is achieved by virtue of the column pruning mechanism of AIRLS, which gradually reduces the size of matrix factors from $m\times 100$ and $n\times 100$ to $m\times 4$ and $n\times 4$, respectively. This way, after only a few initial iterations, when the rank starts to decrease, the per iteration time complexity of AIRLS becomes much smaller than that required in its early iterations, as well as the one of MMMF. 
\subsubsection{MC on Movielens 100K and 10M datasets}
Herein, we focus on testing the performance of AIRLS-MC algorithm on a popular collaborative filtering application i.e. a movie recommender system. To this end, we utilize two well-studied in literature large datasets: the Movielens 100K and the Movielens 10M datasets. Both datasets contain ratings collected over various periods of time by users, with integer 
values ranging from 1-5. Since most of the entries are missing, matrix completion algorithms can be utilized for predicting them. By assuming that there exists a high degree of correlation amongst 
the rating of different users, a low-rank structure can be meaningfully adopted for these datasets. For validation purposes, each of them is splited into two disjoint sets i.e., a training and a test set (the ub.base, ub.test and the ra.train, ra.test are used for the 100K and the 10M dataset, respectively). Note that the 100K dataset contains 100000 ratings of 943 users on 1682 movies with each user having rated at least 20 movies. 
That said, we need to address a quite challenging matrix completion problem, since  93\% of the elements are missing. The situation is even harsher for the 10M dataset, which includes 1 million ratings from 72000 users on 10000 movies and 99\% missing data. The test sets ub.test and ra.test for both datasets contain exactly 10 ratings per user. The state-of-the-art softImpute-ALS algorithm is utilized in this experiment for comparison purposes. Finally, the normalized mean absolute value error (NMAE) defined as $\mathrm{NMAE} = \frac{\sum_{(i,j)\in \Omega}|[\mathbf{U}\mathbf{V}^T]_{ij} - [\mathbf{Y}]_{ij}|}{4\mathrm{card}(\Omega)}$ is used as a performance metric. 

\begin{figure}
\centering
\begin{tabular}{c c}
scenario A & scenario B \\
 \includegraphics[width=0.22\textwidth,height=0.18\textwidth]{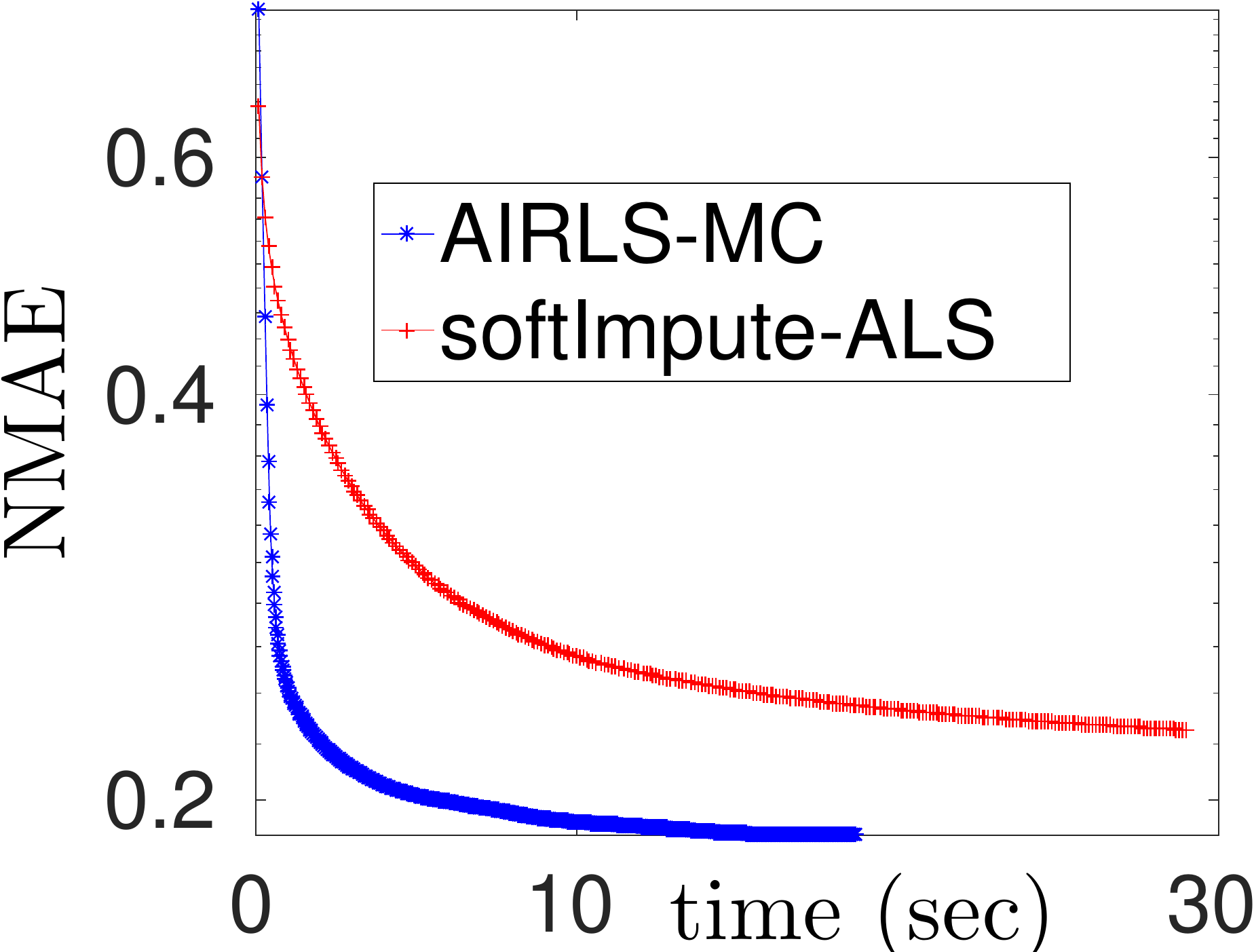} & \includegraphics[width=0.22\textwidth,height=0.18\textwidth]{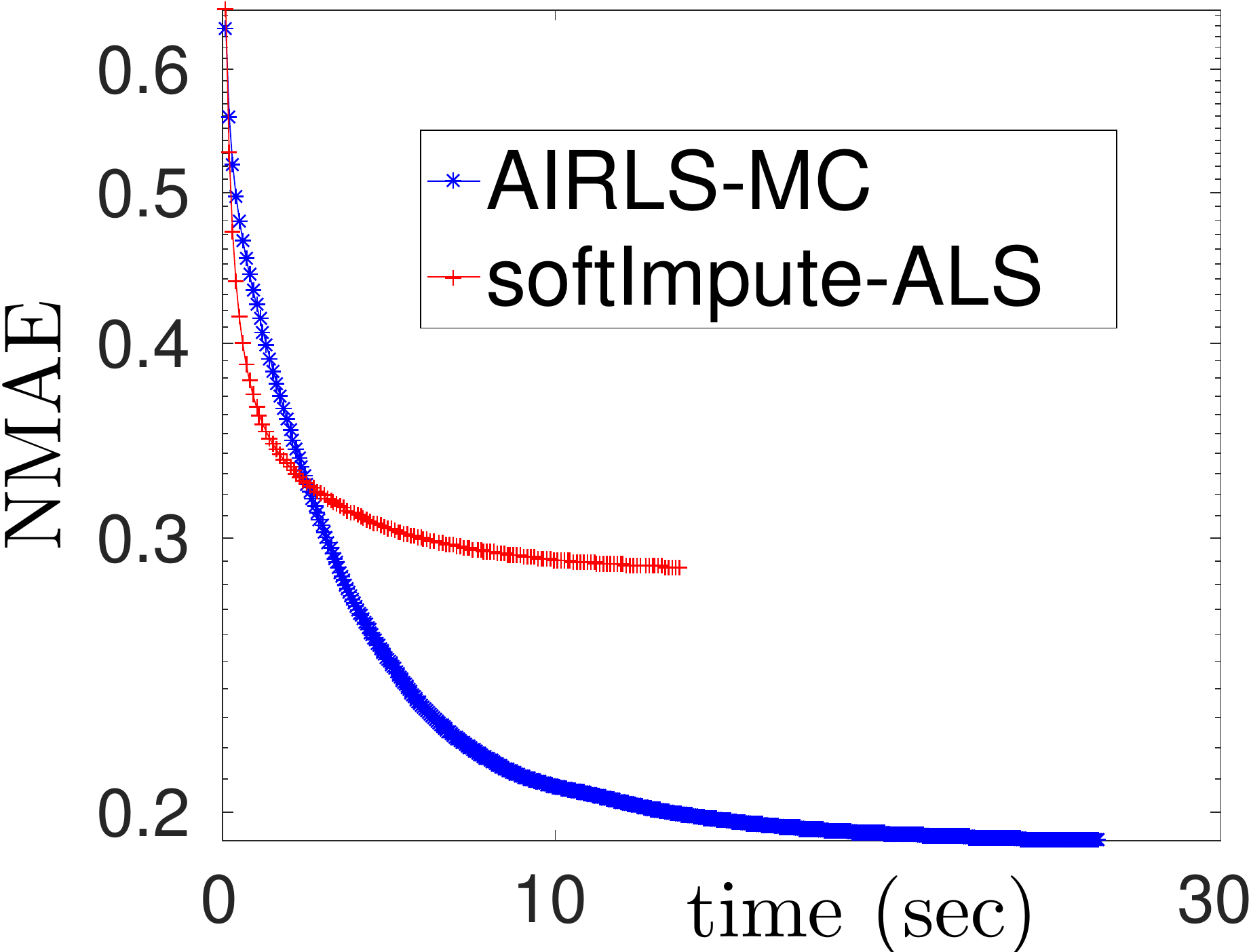}  \\ 
 \includegraphics[width=0.22\textwidth,height=0.18\textwidth]{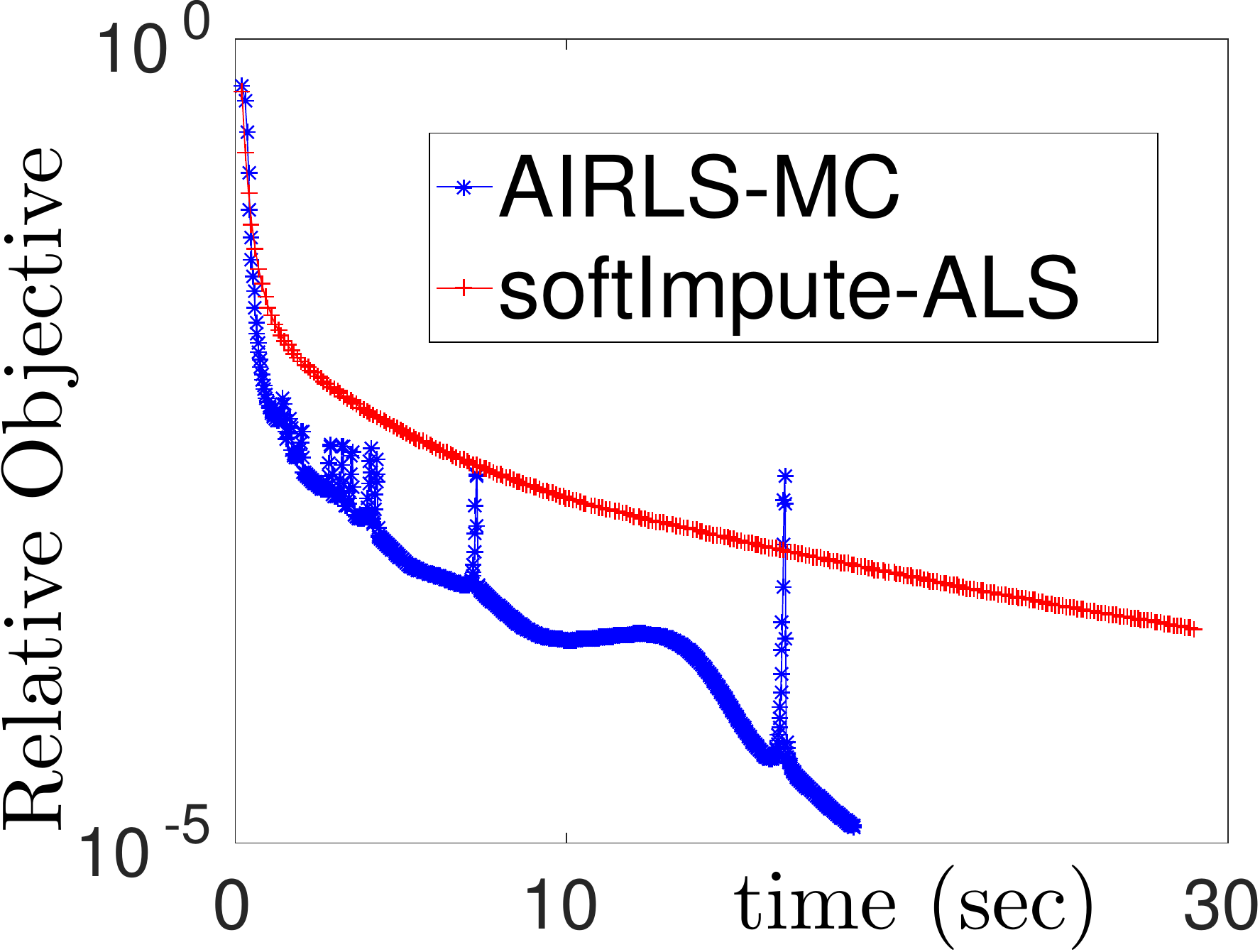} & \includegraphics[width=0.22\textwidth,height=0.18\textwidth]{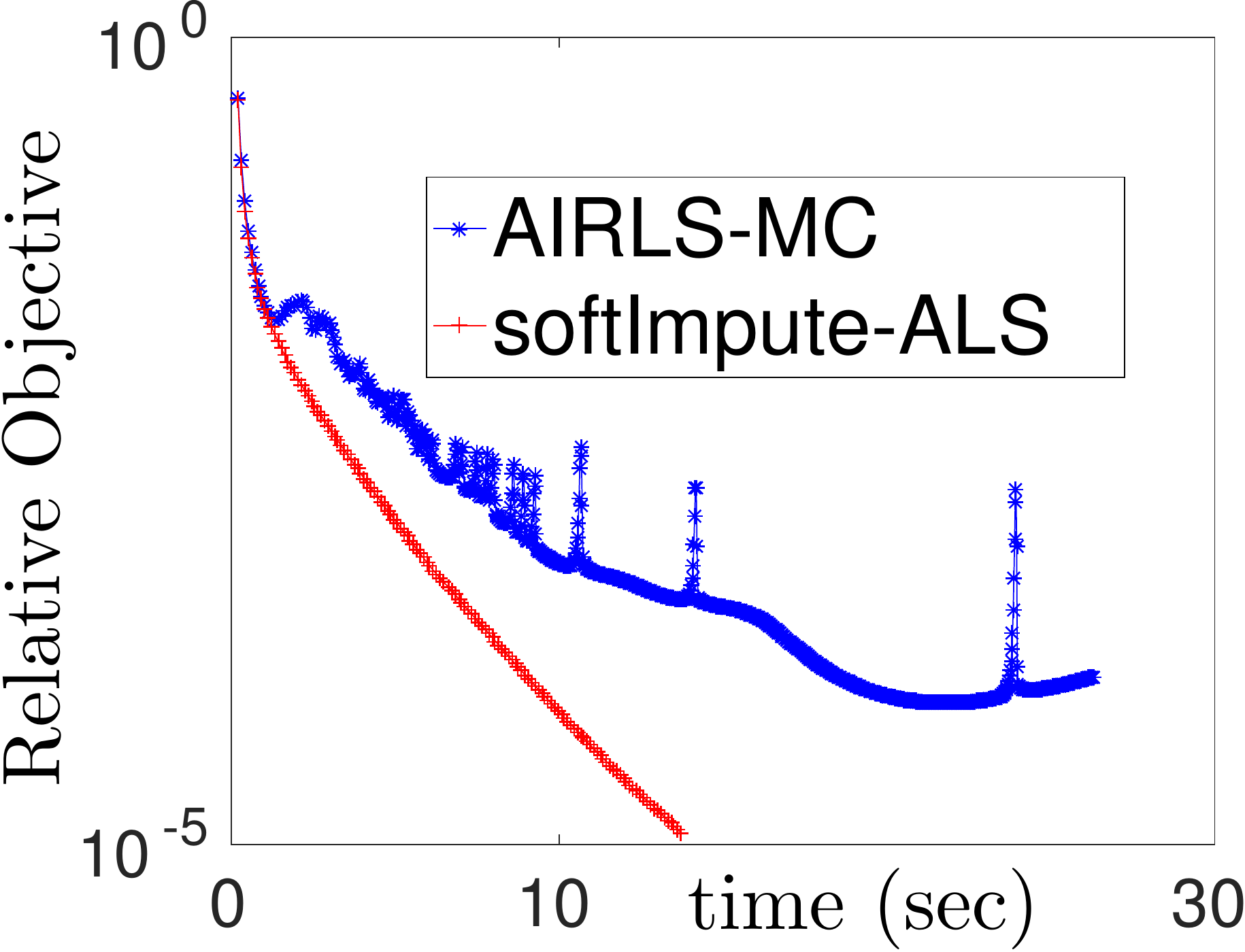} 
 \end{tabular}\vspace{-0.1cm}
 \caption{Evaluation of AIRLS-MC and softImpute-ALS on the Movielens 100K dataset.}
 \vspace{-0.1cm}
 \label{fig:matrix_completion_100K}
\end{figure}

First, we aim at illustrating the behavior of the proposed AIRLS-MC algorithm when it comes to the estimation performance and the speed of convergence. In that vein, for the case of the 100K dataset,
the low-rank promoting parameter $\lambda$ of both AIRLS-MC and softImpute-ALS is selected according to two different scenarios: A) we choose $\lambda$ that achieves the 
minimum NMAE after convergence  and B) we select  $\lambda$ so that the estimated matrices by both the tested algorithms are of the same rank, equal to 10. It should be noted that
the same stopping criterion used in the previous experiment is adopted also here. As it can be seen in Fig. \ref{fig:matrix_completion_100K} and Table \ref{table:movielens100K-results}, 
the proposed AIRLS-MC achieves better performance in terms of the NMAE for both scenarios A and B. The softImpute-ALS algorithm requires less iterations to converge than 
AIRLS-MC. However, the average per-iteration time complexity of AIRLS-MC is significantly less compared to its rival. As is mentioned above, this is attributed to the column pruning 
scheme which decreases to a large degree the computational burden of the algorithm.  This favorable property, results to a much faster convergence of AIRLS-MC as compared to softImpute-ALS in terms of time. It should be noted that in scenario A, the estimated matrices $\hat{\mathbf{U}}$ and $\hat{\mathbf{V}}$ have rank equal to 6. On the other hand, for softImpute-ALS the solution matrices have rank equal to the one used at the initialization stage i.e., 100. In scenario B, softImpute-ALS converged faster than the proposed algorithm. However, this happened at the price of a remarkable deterioration of the NMAE. Lastly, from Fig. \ref{fig:matrix_completion_100K} it can be noticed that the relative objective of AIRLS-MC 
presents abrupt increases at some iterations. It was experimentally verified that those changes (which imply large decreases of the successive values of the objective function) take place at iterations that coincide with zeroings of the columns of the matrix factors. This fact advocates that larger gains are obtained at iterations where the rank is reduced, as we are approaching at the low-rank solution matrices.
\vspace{-0.1cm}
\begin{table}
 \centering
 \resizebox{0.5\textwidth}{!}{
 \begin{tabular}{c  c  c | c | c | c | c |}
 \cline{4-7}
 &  &  &   \small \# Iter & \small msec/iter & total time (sec) & NMAE \\ \cline{1-7}
 \multicolumn{1}{|c}{\multirow{4}{*}{\rotatebox[origin=c]{90}{scenario}}} & \multicolumn{1}{|c|}{\multirow{2}{*}{A}}  &  \multicolumn{1}{c|}{softImpute-ALS }  & 278  & 104,2 & 28,9& 0,2254\\ \cline{3-7}
  \multicolumn{1}{|c}{} &  \multicolumn{1}{|c|}{}  &    \multicolumn{1}{c|}{AIRLS-MC} & 957 & 19,5 & \textbf{18,7} & \textbf{0,1882} \\  \cline{2-7} 
 \multicolumn{1}{|c}{} &\multicolumn{1}{|c|}{\multirow{2}{*}{B}} &  \multicolumn{1}{c|}{softImpute-ALS} &  135 & 101,5 & \textbf{13,7} &  0,2873 \\ \cline{3-7}
 \multicolumn{1}{|c}{} &\multicolumn{1}{|c|}{}& \multicolumn{1}{c|}{AIRLS-MC} &  964  & 27,3 & 26,3 & \textbf{0,1918}\\
   \hline
    \end{tabular}}
    \caption{Results obtained by AIRLS-MC and softImpute-ALS on Movielens 100K dataset.}
    \vspace{-0.8cm}
     \label{table:movielens100K-results}
\end{table}

\begin{figure}
\centering
 \includegraphics[width=0.45\textwidth,height=0.25\textwidth]{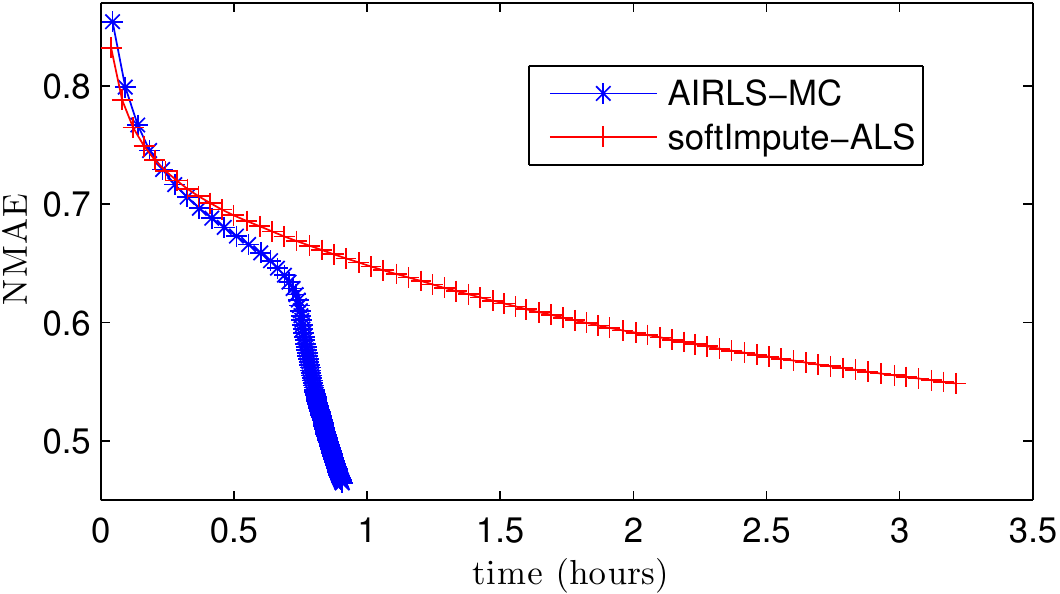} 
 \vspace{-0.1cm}
 \caption{\footnotesize{Evaluation of AIRLS-MC and softImpute-ALS on 10M Movielens dataset.}}
 \vspace{-0.3cm}
 \label{fig:movielens10M}
\end{figure}

Fig. \ref{fig:movielens10M} and Table \ref{table:results10M} show the performance of AIRLS-MC and softImpute-ALS on the 10M Movielens dataset. It should be noted that due to the large scale of this dataset 
the speed of convergence of the algorithms to a descent solution is of crucial importance. The parameter $\lambda$ of AIRLS-MC  is now set to 3000, while for softImpute-ALS $\lambda$ is set, as proposed in \cite{hastie2015matrix}, to 50. The rank is initialized to 100 for both algorithms. In this experiment
the relative tolerance criterion is set to
$10^{-3}$. Interestingly, AIRLS-MC reaches a more accurate solution in terms of the NMAE (evaluated on the test set) in almost 1/3 of the time required by softIMpute-ALS. Again, AIRLS-MC requires more iterations to converge as compared to its competitor. Nevertheless, as it can be easily seen in Fig. \ref{fig:movielens10M}, after  the initial iterations, when the rank starts to decrease and the column pruning mechanism is activated, the time per iteration of AIRLS-MC is dramatically reduced. 
\begin{table}
\resizebox{0.5\textwidth}{!}{
  \begin{tabular}{c |c | c | c| c |}
 \cline{2-5}
    &  \small \# Iter & \small min/iter & total time (min) & NMAE \\ \hline
  \multicolumn{1}{|c|}{ softImpute-ALS} &  71 & 2,71  & 192,6  & 0,5485\\ \hline
   \multicolumn{1}{|c|}{AIRLS-MC} & 134 & 0,40 & 54,4  &  0,4645 \\
   \hline
    \end{tabular}}
    \caption{Results obtained by AIRLS-MC and softImpute-ALS on Movielens 10M dataset.}
    \vspace{-0.5cm}
    \label{table:results10M}
\end{table}
\subsubsection{Music signal decomposition}
\begin{figure}
\centering
 \begin{tabular}{c}
  \includegraphics[width = 0.4\textwidth,height = 0.1\textwidth]{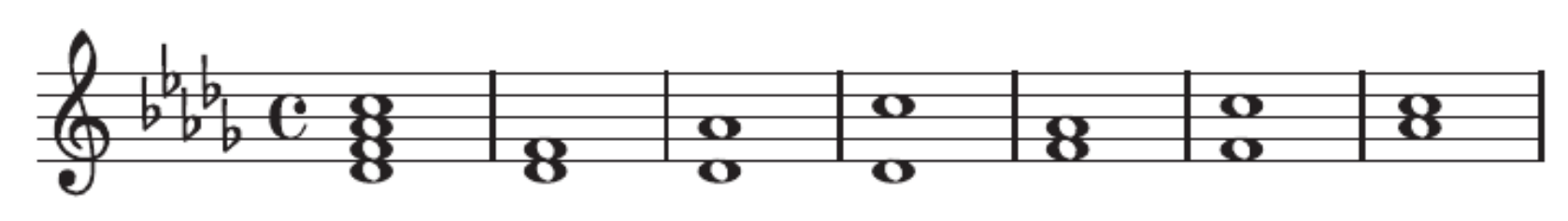} \\
  \includegraphics[width = 0.4\textwidth,height = 0.1\textwidth]{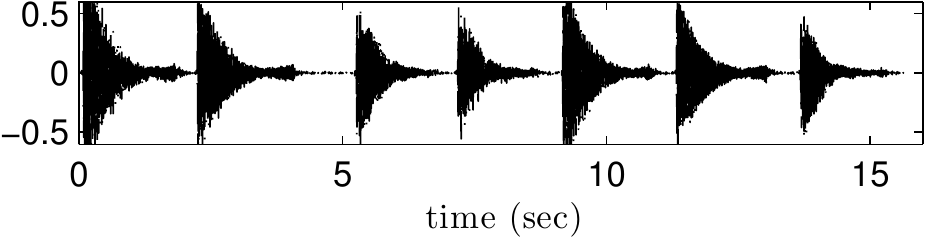} 
 \end{tabular}\vspace{-0.1cm}
 \caption{Music score (top) and original audio signal (bottom)}
 \vspace{-0.1cm}
 \label{fig:piano_seq}
\end{figure}

Herein, we test the competence of AIRLS-NMF algorithm in decomposing a real music signal. For this reason, AIRLS-NMF is compared to the most relevant state-of-the-art algorithm i.e., ARD-NMF. In order to make as much fairer comparisons as possible 
between those two algorithms, the beta function of ARD-NMF algorithm of \cite{tan2013automatic} is reduced to the square Frobenious norm, by appropriately setting the respective parameter. This way, ARD-NMF, likewise to the proposed AIRLS-NMF, is based on  Gaussian i.i.d noise 
assumptions. 
The music signal analyzed, is a short piano sequence i.e., a monophonic 15 seconds-long signal recorded in real conditions, as described in \cite{tan2013automatic}. As it can be noticed
in Fig. \ref{fig:piano_seq}, it is composed of four piano notes that overlap in all the duration thereof. Following the same process as in \cite{tan2013automatic}, the original signal is tranformed into the frequency domain 
via the short-time Fourier transform (STFT). To this end, a Hamming window of size $L=1024$ is utilized. By appropriately setting up the overlapping between the adjacent frames we are led to a spectrogram whereby the signal is represented by 673 frames in 513  frequency bins. The power of this spectrogram is then provided as input to the tested algorithms. The initial rank is set to 20 and the same stopping criterion as in the previous experiments is utilized,
with the threshold in this case set to $10^{-4}$. Moreover, for AIRLS-NMF the parameter setting described in the simulated data experiment is used i.e., we set $\beta_{\mathbf{U}}=\beta_{\mathbf{V}}=10^{-1}$ and $\sigma=10^{-2}$. Finally, the same process described in \cite{tan2013automatic} is followed for reconstructing the music components, i.e., rank one terms of 
the product $\hat{\mathbf{U}}\hat{\mathbf{V}}^T$ in the time domain.

\begin{figure}
\centering
 \begin{tabular}{c c}
  \includegraphics[width = 0.22\textwidth,height = 0.07\textwidth]{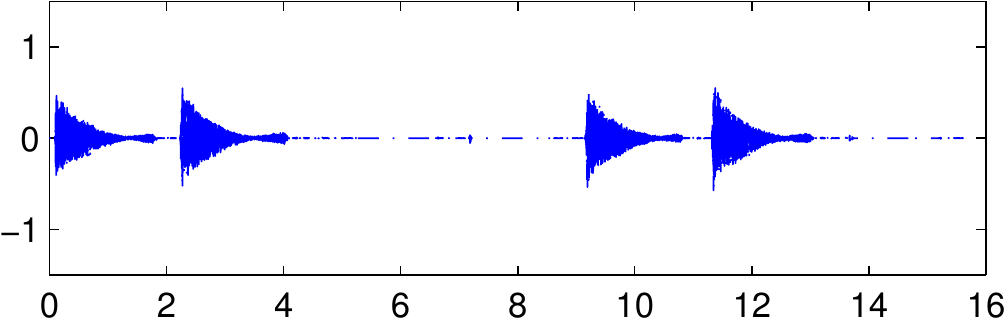} &
  \includegraphics[width = 0.22\textwidth,height = 0.07\textwidth]{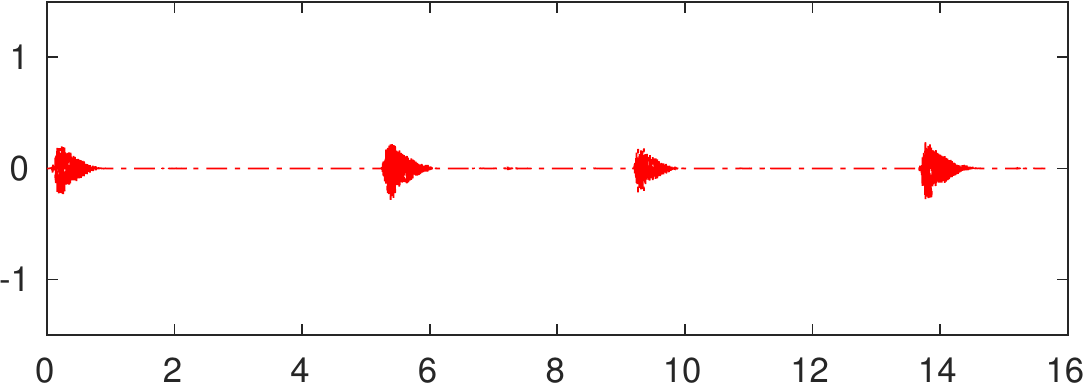}\\
  \includegraphics[width = 0.22\textwidth,height = 0.07\textwidth]{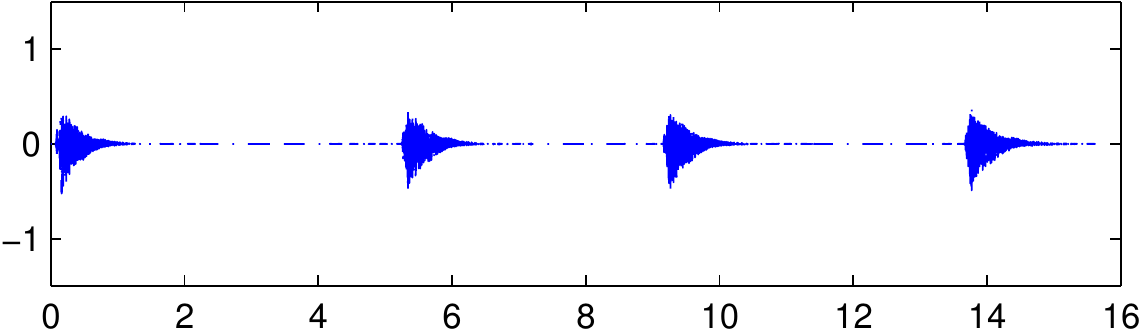} &
  \includegraphics[width = 0.22\textwidth,height = 0.07\textwidth]{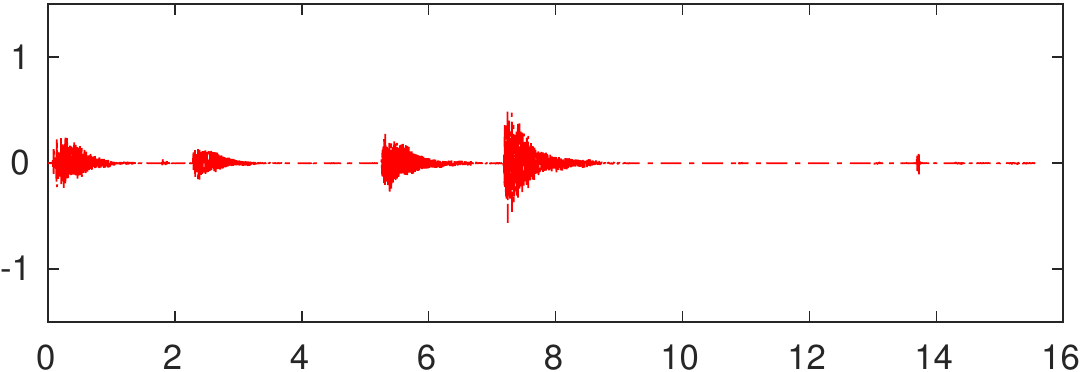} \\
\includegraphics[width = 0.22\textwidth,height = 0.07\textwidth]{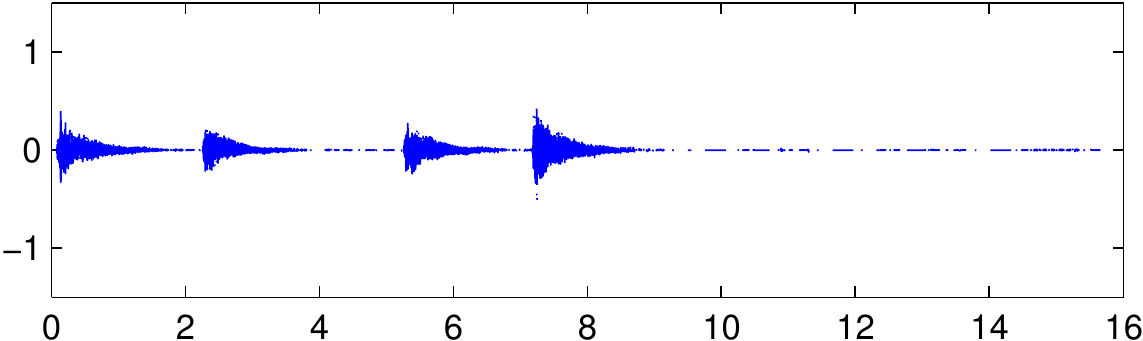} &
\includegraphics[width = 0.22\textwidth,height = 0.07\textwidth]{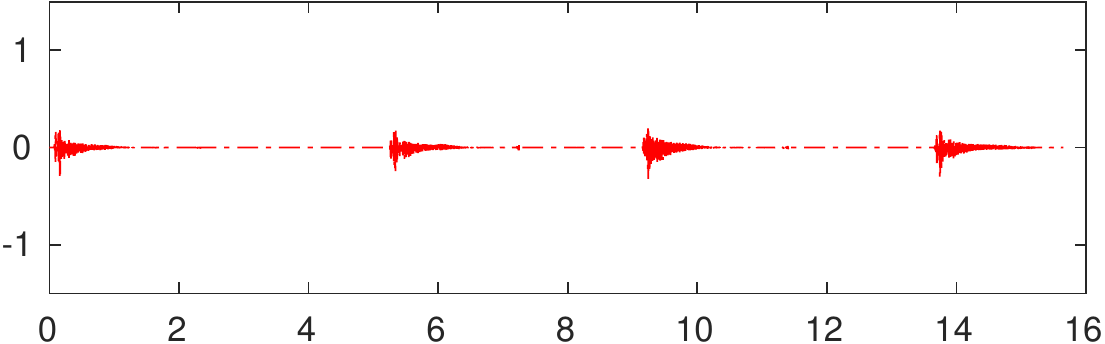} \\
\includegraphics[width = 0.22\textwidth,height = 0.07\textwidth]{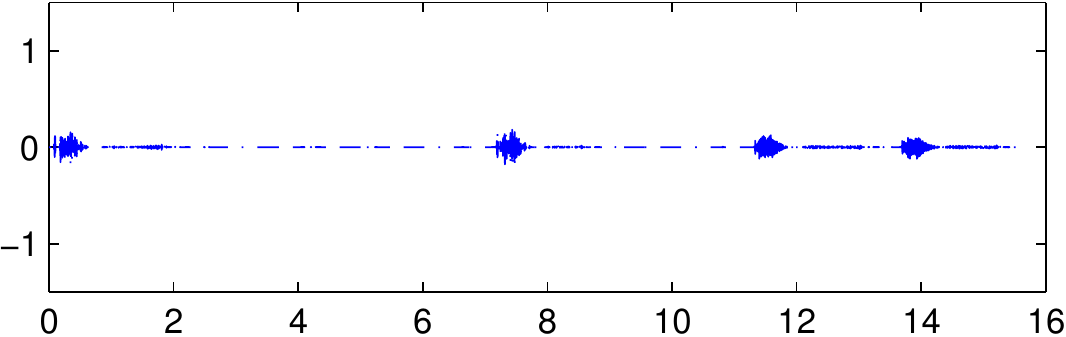} &
\includegraphics[width = 0.22\textwidth,height = 0.07\textwidth]{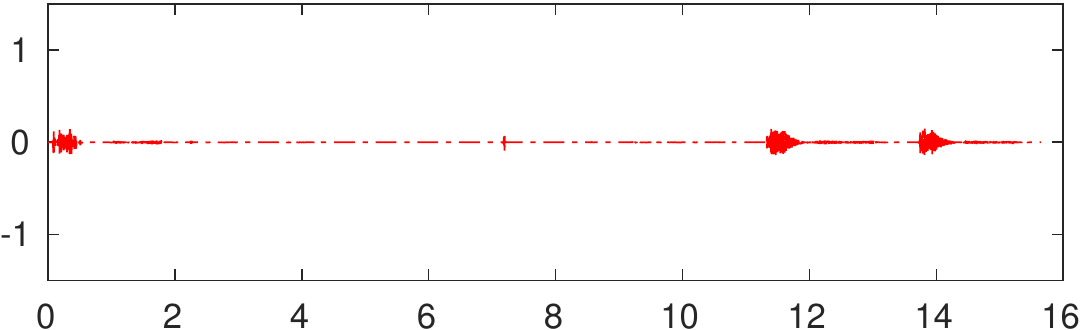} \\
\includegraphics[width = 0.22\textwidth,height = 0.07\textwidth]{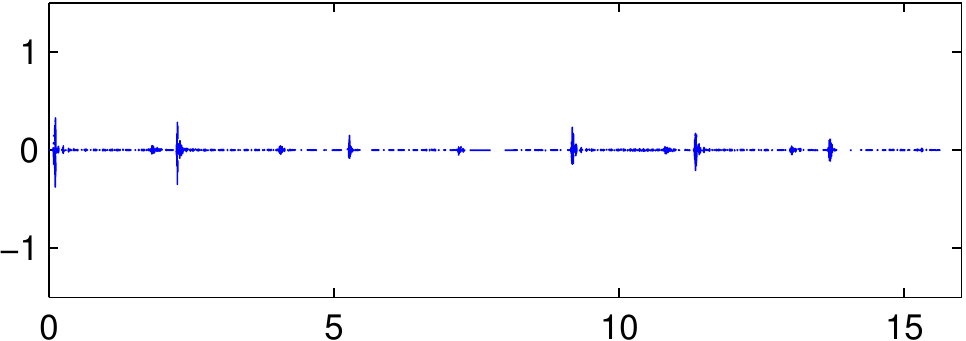} &
\includegraphics[width = 0.22\textwidth,height = 0.07\textwidth]{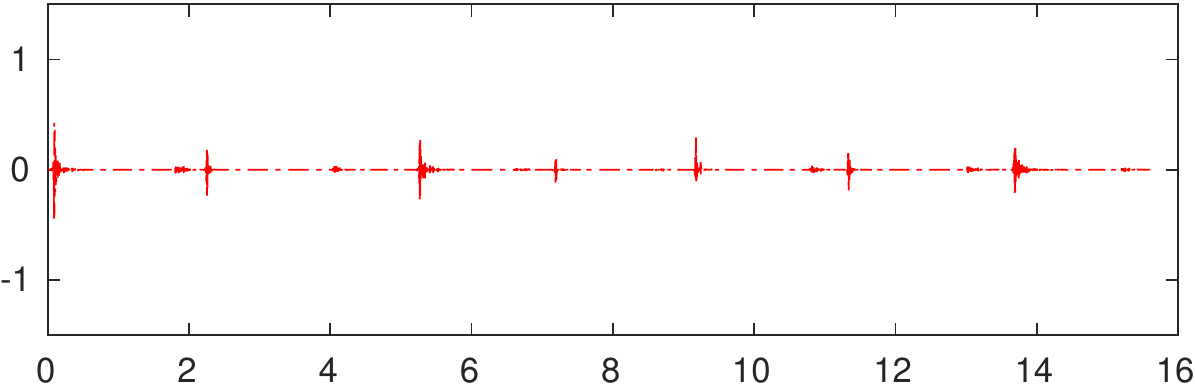} \\
\includegraphics[width = 0.22\textwidth,height = 0.07\textwidth]{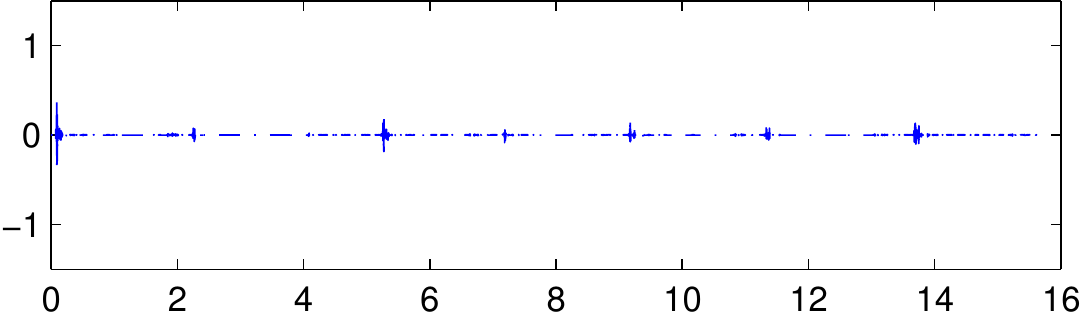} &
  \includegraphics[width = 0.22\textwidth,height = 0.07\textwidth]{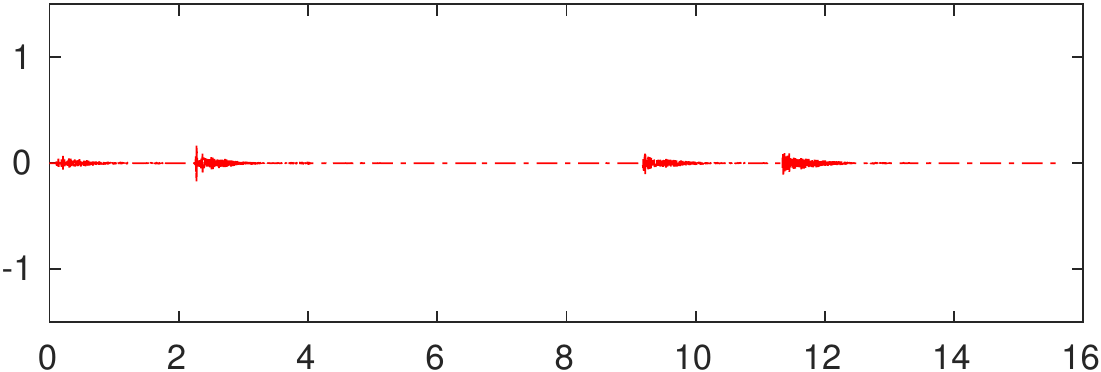} \\
  \includegraphics[width = 0.22\textwidth,height = 0.07\textwidth]{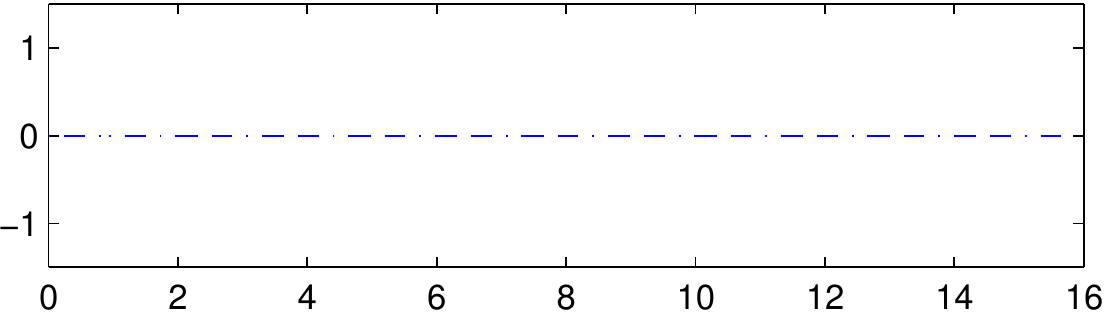} &
  \includegraphics[width = 0.22\textwidth,height = 0.07\textwidth]{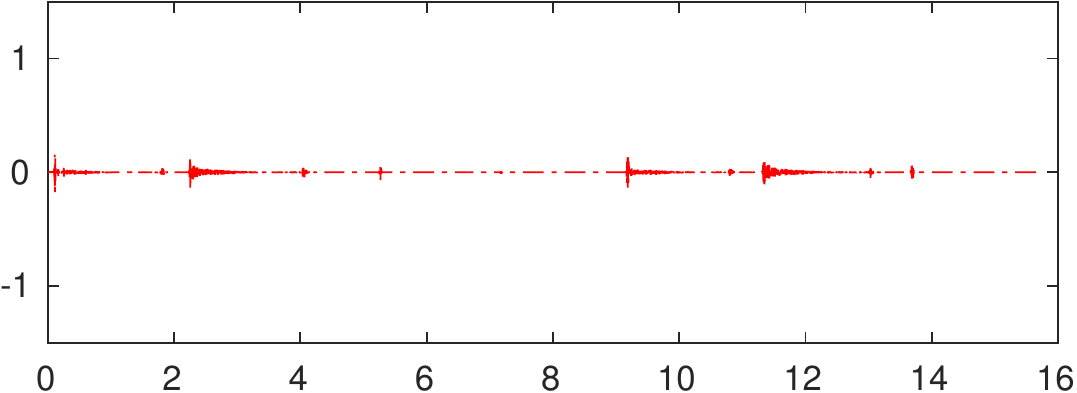} \\
 \includegraphics[width = 0.22\textwidth,height = 0.07\textwidth]{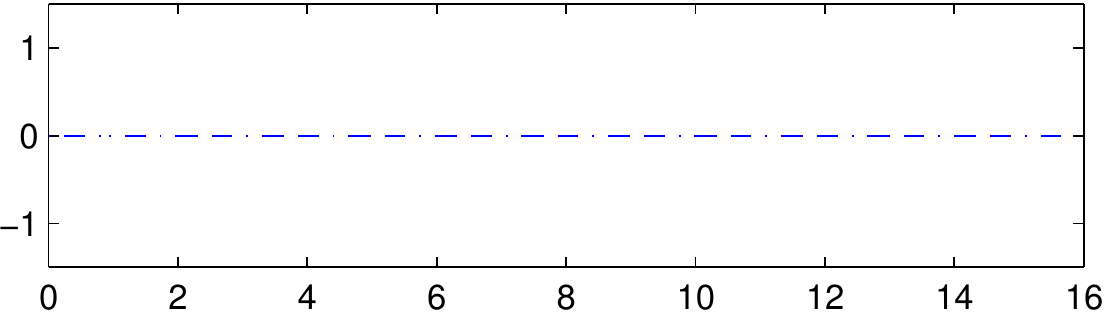} &
  \includegraphics[width = 0.22\textwidth,height = 0.07\textwidth]{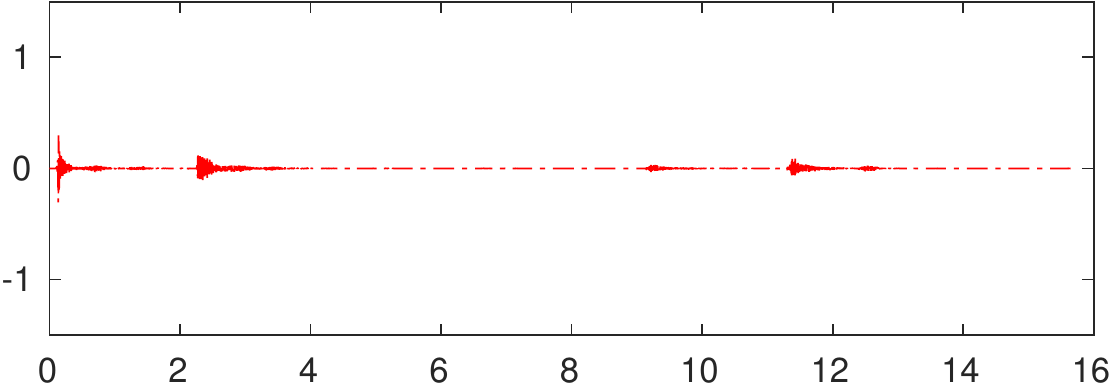} \\
\includegraphics[width = 0.22\textwidth,height = 0.07\textwidth]{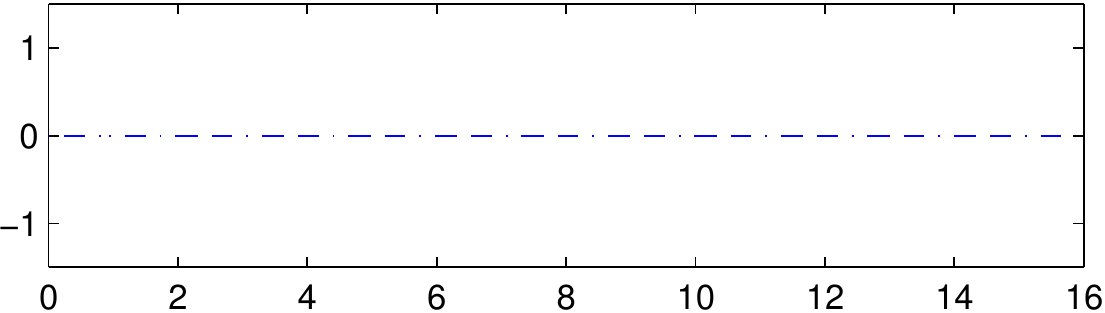} &
\includegraphics[width = 0.22\textwidth,height = 0.07\textwidth]{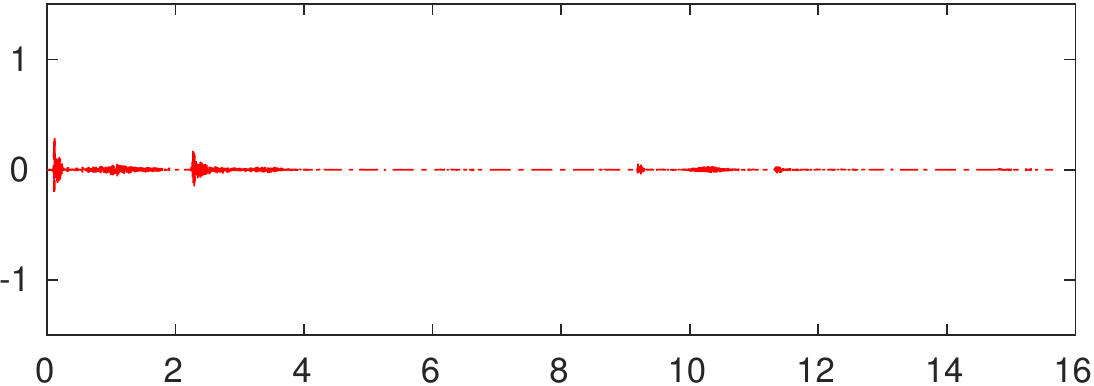} \\
\includegraphics[width = 0.22\textwidth,height = 0.07\textwidth]{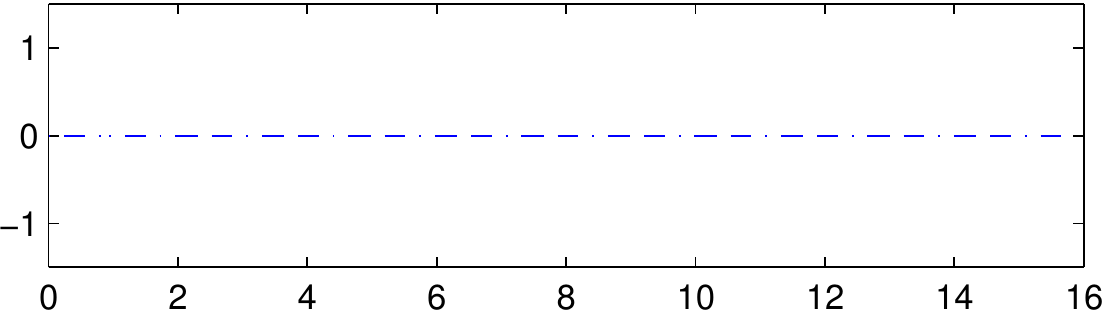} &
\includegraphics[width = 0.22\textwidth,height = 0.07\textwidth]{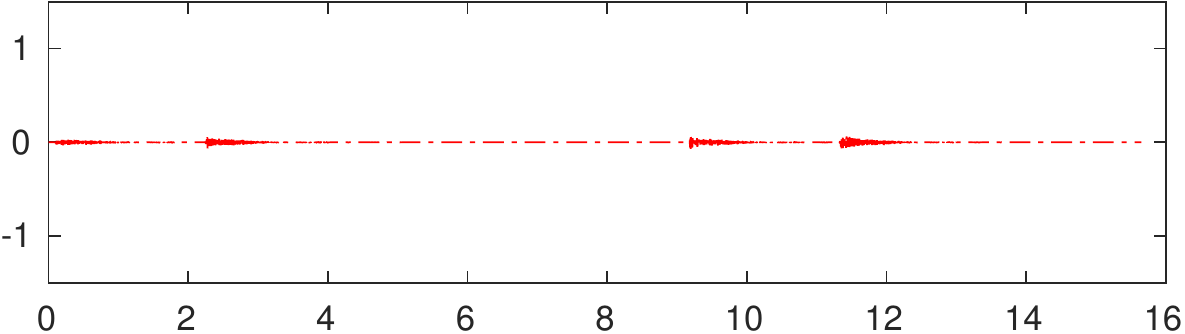} \\
(a) & (b)
 \end{tabular}\vspace{-0.1cm}
 \caption{Music components obtained by (a) AIRLS-NMF  and (b) ARD-NMF  on the short piano sequence.}
 \vspace{-0.0cm}
 \label{fig:results-piano}
\end{figure}

In Fig. \ref{fig:results-piano}, the first 10 components obtained by the two algorithms are ordered in decreasing values of the standard deviations of the time domain waveforms.  
As it can be noticed, AIRLS-NMF estimated the correct number of components, that is 6. Notably, the first four components of AIRLS-NMF correspond to the four notes while the rest
two ones come from the sound of a hammer hitting the strings and the sound produced by the sustain pedal when it is released. On the contrary, ARD-NMF estimated 20 components, meaning that no rank minimization took place thus implying a data overfitting behavior.  It should be emphasized that the favorable performance of AIRLS-NMF occurs though the noise 
is implicitly modeled as Gaussian i.i.d. Interestingly, as it 
can be seen in \cite{tan2013automatic}, AIRLS-NMF performed similarly to ARD IS-NMF, i.e., the version of ARD-NMF which makes more appropriate assumptions as to the noise statistics, by modeling it as Itakura-Saito.   
\vspace{-0.4cm}
\vspace{-0.0cm}
\section{Conclusion}
This paper presents a novel generic formulation of the low-rank matrix factorization problem. Borrowing ideas from iteratively reweighted approaches for rank minimization, a reweighted version of the sum of the squared Frobenious norms of the matrix factors i.e., a non-convex variational characterization of the nuclear norm, is defined. The proposed framework encapsulates other state-of-the-art approaches for low-rank imposition on the matrix factorization setting. By focusing on a specific instance of this scheme we generate a joint-column sparsity inducing regularizer that couples the columns of the matrix factors. The ubiquity of the proposed approach is demonstrated in the problems of denoising, matrix completion and nonnegative matrix factorization (NMF). To this end, under the block successive upper bound minimization (BSUM) framework, Newton-type algorithms are devised for addressing the afore-mentioned problems. The efficiency of the proposed algorithms in handling big and high-dimensional data as compared to other state-of-the-art algorithms is illustrated in a wealth of simulated and real data experiments.

%
%
\bibliographystyle{IEEEtran}
\bibliography{IEEEabrv,refs_report}
\setcounter{page}{1}
\section*{Appendix}
\subsection*{Proof of Lemma 1}
In denoising and matrix completion, the surrogate functions $l(\mathbf{U}| \mathbf{U}_k,\mathbf{V}_k)$ and $g(\mathbf{V}|\mathbf{U}_{k+1},\mathbf{V}_k)$ given in eqs. (\ref{eq:upper_bound_l}) and (\ref{eq:upper_bound_g}), are twice continuously differentiable and constitute approximations of the second order Taylor expansions of the initial cost functions around ($\mathbf{U}_k,\mathbf{V}_k$) and ($\mathbf{U}_{k+1},\mathbf{V}_k$) respectively. 
In (\ref{eq:upper_bound_l}), the true Hessian $\mathbf{H}_{\mathbf{U}_k}$ of $f(\mathbf{U},\mathbf{V}_k)$ at $\mathbf{U}_k$ has been approximated by the $md \times md$ positive-definite block diagonal matrix $\bar{\mathbf{H}}_{\mathbf{U}_k}$ defined in (\ref{hessian_bd}).
$\bar{\mathbf{H}}_{\mathbf{V}_k}$ is similarly defined.
Our analysis is next focused on $l(\mathbf{U}|\mathbf{U}_k,\mathbf{V}_k)$. It can be easily shown that similar derivations can be made for $g(\mathbf{V}|\mathbf{U}_{k+1},\mathbf{V}_{k})$.
As it can be seen by eq. (\ref{eq:upper_bound_l}), $l(\mathbf{U}|\mathbf{U}_k,\mathbf{V}_k)$ equals $f(\mathbf{U},\mathbf{V}_k)$ at $(\mathbf{U}_k,\mathbf{V}_k)$. In order to show that it majorizes $f(\mathbf{U},\mathbf{V}_k)$ for all other points closeby, it suffices to show that matrix $\mathbf{A} = \bar{\mathbf{H}}_{\mathbf{U}_k} - \mathbf{H}_{\mathbf{U}_k}$ is positive semi-definite \cite{razaviyayn2013unified}. Next we prove that for each of the two problems examined, the above-mentioned property holds for $\mathbf{A}$.

In denoising $\tilde{\mathbf{H}}_{\mathbf{U}_k} = \mathbf{V}_k^T\mathbf{V}_k + \lambda\mathbf{D}_{(\mathbf{U}_k,\mathbf{V}_k)}$, where $\mathbf{D}_{(\mathbf{U}_k,\mathbf{V}_k)}$ is defined in eq. (\ref{definition_D}). Moreover for the exact Hessian $\mathbf{H}_{\mathbf{U}_k}$ at $\mathbf{U}_k$ we have 
\small
 \begin{align}
 &{\mathbf{H}}_{\mathbf{U}_k} =  \nonumber \\
 &\left[ \begin{array}{c c c c} 
            \mathbf{V}_k^T\mathbf{V}_k+ \mathbf{K}_{11} & \mathbf{K}_{12} &  \dots & \mathbf{K}_{1m} \\                       
              \mathbf{K}_{12} &    \mathbf{V}^T_k\mathbf{V}_k+ \mathbf{K}_{22}  & \ddots & \vdots \\      
              \vdots    & \ddots & \ddots &  \mathbf{K}_{(m-1)m} \\
              \mathbf{K}_{1m} & \dots &  \mathbf{K}_{(m-1)m} & \mathbf{V}^T_k\mathbf{V}_k+ \mathbf{K}_{mm}
              \end{array} \right] \label{exact_Hessian_denoising}
\end{align}
\normalsize
where 
\small
\begin{align}
&\mathbf{K}_{ij} = \nonumber \\
&\begin{cases} \mathrm{diag}\left(\frac{\|\boldsymbol{\mathit{u}}^k_1\|^2_2  + \|\boldsymbol{\mathit{v}}^k_1\|^2_2 - (u_{i1}^k)^2+ \eta^2}{\left(\|\boldsymbol{\mathit{u}}^k_1\|^2_2  + \|\boldsymbol{\mathit{v}}^k_1\|^2_2 + \eta^2\right)^{\frac{3}{2}}}, \cdots, \frac{\|\boldsymbol{\mathit{u}}^k_d\|^2_2  + \|\boldsymbol{\mathit{v}}^k_d\|^2_2 - (u_{id}^k)^2+ \eta^2}{\left(\|\boldsymbol{\mathit{u}}^k_d\|^2_2  + \|\boldsymbol{\mathit{v}}^k_d\|^2_2 + \eta^2\right)^{\frac{3}{2}}} \right), \text{if} \;  i=j \\ 
\\ \mathrm{diag}\left(\frac{-u_{i1}^k u_{j1}^k}{\left(\|\boldsymbol{\mathit{u}}^k_1\|^2_2  + \|\boldsymbol{\mathit{v}}^k_1\|^2_2 + \eta^2\right)^{\frac{3}{2}}}, \cdots, \frac{-u_{id}^k u_{jd}^k}{\left(\|\boldsymbol{\mathit{u}}^k_d\|^2_2  + \|\boldsymbol{\mathit{v}}^k_d\|^2_2 + \eta^2\right)^{\frac{3}{2}}} \right), \text{if} \; i \neq j \end{cases} \label{kij}
\end{align}
\normalsize
Hence matrix $\mathbf{A}$ takes the form given at the top of the next page.  
\tiny
\begin{figure*}
\begin{align}
 & {\mathbf{A}}  = \nonumber \\
 &\left[ \begin{array}{c c c c} 
                \mathbf{D}_{(\mathbf{U}_k,\mathbf{V}_k)} - \mathbf{K}_{11} & -\mathbf{K}_{12} &  \dots & -\mathbf{K}_{1m} \\                       
              -\mathbf{K}_{12} &   \mathbf{D}_{(\mathbf{U}_k,\mathbf{V}_k)} - \mathbf{K}_{22}  & \ddots & \vdots \\      
              \vdots    & \ddots & \ddots &  -\mathbf{K}_{(m-1)m} \\
             - \mathbf{K}_{1m} & \dots &  -\mathbf{K}_{(m-1)m}& \mathbf{D}_{(\mathbf{U}_k,\mathbf{V}_k)} - \mathbf{K}_{mm}
              \end{array} \right] \label{eq:Amatrix} 
 &\equiv  \left[ \begin{array}{c c c c} 
                \mathbf{A}_{11}  & \mathbf{A}_{12} &  \dots & \mathbf{A}_{1m} \\                       
              \mathbf{A}_{12} &  \mathbf{A}_{22}  & \ddots & \vdots \\      
              \vdots    & \ddots & \ddots &  \mathbf{A}_{(m-1)m} \\
              \mathbf{A}_{1m} & \dots &  \mathbf{A}_{(m-1)m} &  \mathbf{A}_{mm}
              \end{array} \right].
 \end{align}
 \end{figure*}
 \normalsize
 Elaborating on $\mathbf{A}$ we get from (\ref{eq:Amatrix}), (\ref{kij}) and (\ref{definition_D}), 
\begin{align}
 \mathbf{A}_{ij} =  \mathrm{diag}\Big(  
           \frac{u^k_{i1}u^k_{j1}}{\left(\|\boldsymbol{\mathit{u}}^k_1\|^2_2  + \|\boldsymbol{\mathit{v}}^k_1\|^2_2 + \eta^2\right)^{\frac{3}{2}}}, \cdots,\nonumber \\
           \frac{u^k_{id}u^k_{jd}}{\left(\|\boldsymbol{\mathit{u}}^k_d\|^2_2  + \|\boldsymbol{\mathit{v}}^k_d\|^2_2 + \eta^2\right)^{\frac{3}{2}}}\Big).\label{Aij}                      
\end{align}
Notice that for \\
$\mathbf{B}_i = \mathrm{diag}\left(\frac{u^k_{i1}}{\left(\|\boldsymbol{\mathit{u}}^k_1\|^2_2  + \|\boldsymbol{\mathit{v}}^k_1\|^2_2 + \eta^2\right)^{\frac{3}{4}}},\dots,\frac{u^k_{id}}{\left(\|\boldsymbol{\mathit{u}}^k_d\|^2_2  + \|\boldsymbol{\mathit{v}}^k_d\|^2_2 + \eta^2\right)^{\frac{3}{4}}}\right)$, $\mathbf{A}_{ij}=\mathbf{B}_i^T\mathbf{B}_j$. So by defining $\mathbf{B}=[\mathbf{B}_1,\ldots,\mathbf{B}_d]$, it is straightforward that $\mathbf{A}=\mathbf{B}^T\mathbf{B}$, that is $\mathbf{A}$ is positive semi-definite. 

In matrix completion, the exact Hessian $\mathbf{H}_{\mathbf{U}_k}$ differs from that given in (\ref{exact_Hessian_denoising}) in the diagonal blocks only. More specifically, the $i$th diagonal block of $\mathbf{H}_{\mathbf{U}_k}$ takes now the form $\mathbf{V}^T\boldsymbol{\Phi}_i\mathbf{V} + \mathbf{K}_{ii}$, where $\boldsymbol{\Phi}_i$ is a $n \times n$ diagonal matrix containing ones on indexes included in the set $\Omega$ and related to the $i$th row of $\mathbf{Y} $ and zeros elsewhere. Since $\mathbf{V}^T\mathbf{V}- (\mathbf{V}^T\boldsymbol{\Phi}_i\mathbf{V})\succeq 0$, we can easily follow the same path as above for proving the semi-definiteness of the respective matrix $\mathbf{A}$. 
\subsection*{Proof of Lemma 2}
Working as in the proof of Lemma 1, it can be shown that the surrogate functions are upper bounds of the actual cost functions, if matrices $\frac{1}{a^k_{\mathbf{U}}}\tilde{\mathbf{H}}^{\mathcal{I}^k_{\mathbf{U}}}_{\mathbf{U}} - \mathbf{H}_{\mathbf{U}_k}$ and $\frac{1}{a^k_{\mathbf{V}}}\tilde{\mathbf{H}}^{\mathcal{I}^k_{\mathbf{V}}}_{\mathbf{V}} - \mathbf{H}_{\mathbf{V}_k}$ are positive semi-definite. By using inequalities in the form of $\lambda_{min}(\mathbf{A})\|\mathbf{x}\|^2_2\leq \|\mathbf{Ax}\|^2_2\leq \lambda_{max}(\mathbf{A})\|\mathbf{x}\|^2_2 $ (where $\lambda_{min}(\mathbf{A})$ and $\lambda_{max}(\mathbf{A})$ denote the minimum and the maximum eigenvalues of matrix $\mathbf{A}$, respectively) it can be easily verified that this property holds always, if $a^k_{\mathbf{U}}$ and $a^k_{\mathbf{V}}$ are bounded above as stated in the Lemma.
\subsection*{Proof of Proposition 2}
The following analysis is the same for the denoising and matrix completion problems. From Lemma 1 we have,
 \begin{align}
  l(\mathbf{U}|\mathbf{U}_k,\mathbf{V}_k) \geq f(\mathbf{U},\mathbf{V}_k)
 \end{align}
 Since $\mathbf{U}_{k+1} = \underset{\mathbf{U}}{\mathrm{arg min}}\;\;\; l(\mathbf{U}|\mathbf{U}_k,\mathbf{V}_k) $ we get 
 \begin{align}
    l(\mathbf{U}_{k+1}|\mathbf{U}_k,\mathbf{V}_k) \leq l(\mathbf{U}_k|\mathbf{U}_k,\mathbf{V}_k) \equiv f(\mathbf{U}_k,\mathbf{V}_k)
    \label{eq:prop_1_c}
 \end{align}
 and hence 
 \begin{align}
  f(\mathbf{U}_{k+1},\mathbf{V}_k) \leq f(\mathbf{U}_k,\mathbf{V}_k).
  \label{eq:prop_1_a}
 \end{align}
Following the same rationale, and since $\mathbf{V}_{k+1} = \underset{\mathbf{V}}{\mathrm{arg min}}\;\;\; g(\mathbf{V}|\mathbf{U}_{k+1},\mathbf{V}_k)$ we get
\begin{align}
     g(\mathbf{V}_{k}|\mathbf{U}_{k+1},\mathbf{V}_k) \equiv f(\mathbf{U}_{k+1},\mathbf{V}_k) \geq \nonumber \\
     g(\mathbf{V}_{k+1}|\mathbf{U}_{k+1},\mathbf{V}_k) \geq f(\mathbf{U}_{k+1},\mathbf{V}_{k+1}) 
     \label{eq:prop_1_b}
\end{align}
Combining (\ref{eq:prop_1_a}) and (\ref{eq:prop_1_b}) we get (\ref{eq:prop_1}).
 
In nonnegative matrix factorization, by invoking Proposition 2.4.1 of \cite{bertsekas1999nonlinear}, we have that there exist an $\bar{a}_\mathbf{U}$ which guarantees that for every 
 $a^k_{\mathbf{U}}\in (0,\bar{a}_{\mathbf{U}})$ we have 
 \begin{align}
  f(\mathbf{U}_{k+1}(a^k_{\mathbf{U}}),\mathbf{V}_k) \leq f(\mathbf{U}_k,\mathbf{V}_k)
  \label{eq:prop_2_a}
 \end{align}
  Similarly, there exists $a^k_{\mathbf{V}}\in (0,\bar{a}_{\mathbf{V}})$ for which 
  \begin{align}
  f(\mathbf{U}_{k+1}(a^k_{\mathbf{U}}),\mathbf{V}_{k+1}(a^k_{\mathbf{V}})) \leq f(\mathbf{U}_{k+1}(a^k_{\mathbf{U}}),\mathbf{V}_{k}))
  \label{eq:prop_2_b}
 \end{align}
  Relations (\ref{eq:prop_2_a}) and (\ref{eq:prop_2_b}) lead us to (\ref{eq:prop_1}). 
 \subsection*{Proof of Lemma 3}
 Using Lemma 1, we have:
\begin{enumerate}
\item For Algorithms 1,2: \\[-0.1cm]
\begin{align}
 f(\mathbf{U}_k,\mathbf{V}_k) - f(\mathbf{U}_{k+1},\mathbf{V}_k) \geq \nonumber \\
 l(\mathbf{U}_k|\mathbf{U}_k,\mathbf{V}_k) - l(\mathbf{U}_{k+1}|\mathbf{U}_k,\mathbf{V}_k) \;\;\; \text{and} \label{eq_lemma_2_1}\\
 f(\mathbf{U}_{k+1},\mathbf{V}_k) - f(\mathbf{U}_{k+1},\mathbf{V}_{k+1}) \geq \nonumber \\
 g(\mathbf{V}_k|\mathbf{U}_{k+1},\mathbf{V}_k) - g(\mathbf{V}_{k+1}|\mathbf{U}_{k+1},\mathbf{V}_k) \label{eq_lemma_2_2}
\end{align}
Adding (\ref{eq_lemma_2_1}) and (\ref{eq_lemma_2_2}) we reach to the following inequality
\begin{align}
 &f(\mathbf{U}_k,\mathbf{V}_k) - f(\mathbf{U}_{k+1},\mathbf{V}_{k+1})  \geq \nonumber \\
 &l(\mathbf{U}_k|\mathbf{U}_k,\mathbf{V}_k) - l(\mathbf{U}_{k+1}|\mathbf{U}_k,\mathbf{V}_k) \nonumber \\
 &  + g(\mathbf{V}_k|\mathbf{U}_{k+1},\mathbf{V}_k) - g(\mathbf{V}_{k+1}|\mathbf{U}_{k+1},\mathbf{V}_k) \label{eq_lemma_2_3}
\end{align}
Since $\mathbf{U}_{k+1}$ and $\mathbf{V}_{k+1}$ are stationary points of $l(\mathbf{U}|\mathbf{U}_k,\mathbf{V}_k)$ and $g(\mathbf{V}| \mathbf{U}_{k+1},\mathbf{V}_k)$  respectively\\ ($\nabla_{\mathbf{U}}l(\mathbf{U}_{k+1}|\mathbf{U}_k,\mathbf{V}_k)= \mathbf{0}$ and $\nabla_{\mathbf{V}}g(\mathbf{V}_{k+1}|\mathbf{U}_{k+1},\mathbf{V}_k)=\mathbf{0}$) and by their second order Taylor expansions around $(\mathbf{U}_{k+1},\mathbf{V}_k)$ and $(\mathbf{U}_{k+1},\mathbf{V}_{k+1})$ we have 
\begin{align}
 & l(\mathbf{U}_k|\mathbf{U}_k,\mathbf{V}_k) - l(\mathbf{U}_{k+1}| \mathbf{U}_k,\mathbf{V}_k) = \nonumber \\
 & \frac{1}{2}\mathrm{tr}\{\left(\mathbf{U}_k - \mathbf{U}_{k+1}\right)\big(\mathbf{V}^T_k\mathbf{V}_k + \nonumber \\
 &\lambda\mathbf{D}_{(\mathbf{U}_k,\mathbf{V}_k)}\big)\left(\mathbf{U}_k - \mathbf{U}_{k+1}\right)^T \} \\
 &= \frac{1}{2}\|\mathbf{V}_k\left(\mathbf{U}_k - \mathbf{U}_{k+1}\right)^T \|_F^2  + \nonumber \\
 &\frac{\lambda}{2}\|\mathbf{D}^{\frac{1}{2}}_{(\mathbf{U}_k,\mathbf{V}_k)}\left(\mathbf{U}_k - \mathbf{U}_{k+1}\right)^T \|^2_F \label{eq_lemma_2_4}
 \end{align}
and
\begin{align}
 &g(\mathbf{V}_k|\mathbf{U}_{k+1},\mathbf{V}_k) - g(\mathbf{V}_{k+1}| \mathbf{U}_{k+1},\mathbf{V}_k) = \nonumber \\
 &\frac{1}{2}\mathrm{tr}\{\left(\mathbf{V}_k - \mathbf{V}_{k+1}\right)\big(\mathbf{U}^T_{k+1}\mathbf{U}_{k+1} +  \nonumber \\
 &\lambda\mathbf{D}_{(\mathbf{U}_{k+1},\mathbf{V}_{k})}\big)\left(\mathbf{V}_{k+1} - \mathbf{V}_{k}\right)^T \} \\
 &= \frac{1}{2}\|\mathbf{U}_{k+1}\left(\mathbf{V}_k - \mathbf{V}_{k+1}\right)^T \|_F^2  + \nonumber \\
 &\frac{\lambda}{2}\|\mathbf{D}^{\frac{1}{2}}_{(\mathbf{U}_{k+1},\mathbf{V}_k)}\left(\mathbf{V}_k - \mathbf{V}_{k+1}\right)^T \|^2_F \label{eq_lemma_2_5}
 \end{align}
 Combining (\ref{eq_lemma_2_4}), (\ref{eq_lemma_2_5}) and (\ref{eq_lemma_2_3}) we get inequality (\ref{lemma_2_a}).
 \item For Algorithm 3: \\
 Inequality (\ref{lemma_2_b}) can be  derived following a similar process as above. However there exist two subtle points which lead us to a slightly different  lower bound compared to that of (\ref{lemma_2_a}). More concretely, the first part of $\Delta^b((\mathbf{U}_k,\mathbf{V}_k),(\mathbf{U}_{k+1},\mathbf{V}_{k+1})) $ is now determined by the approximate Hessian adopted for the NMF problem. Second, the constrained nature of the optimization problem is translated into the modified condition of stationarity, which results to the inclusion of two additional positive terms i.e., $\mathrm{tr}\{(\mathbf{U}_k-\mathbf{U}_{k+1})\nabla_{\mathbf{U}}f(\mathbf{U}_k,\mathbf{V}_k)$  and $\mathrm{tr}\{(\mathbf{V}_{k}-\mathbf{V}_{k+1})\nabla_{\mathbf{V}}f(\mathbf{U}_{k+1},\mathbf{V}_k)$. 
\end{enumerate}
\subsection*{Proof of Lemma 4}
If $(\mathbf{U},\mathbf{V})$ is a fixed point, i.e. $\mathbf{U} = \mathbf{U}_{\ast}$ and $\mathbf{V}=\mathbf{V}_{\ast}$, then it is easily shown that $\Delta^a((\mathbf{U},\mathbf{V}),(\mathbf{U}_{\ast},\mathbf{V}_{\ast}))=0$ and $\Delta^b((\mathbf{U},\mathbf{V}),(\mathbf{U}_{\ast},\mathbf{V}_{\ast}))=0$. Conversely, using  (\ref{eq_lemma_2_4}) and (\ref{eq_lemma_2_5}) and since all the summands of $\Delta^a((\mathbf{U},\mathbf{V}),(\mathbf{U}_{\ast},\mathbf{V}_{\ast}))$ are positive, we have that if $\Delta^a((\mathbf{U},\mathbf{V}),(\mathbf{U}_{\ast},\mathbf{V}_{\ast}))=0$ then 
\begin{align}
 l(\mathbf{U}|\mathbf{U},\mathbf{V}) - l(\mathbf{U}_{\ast}| \mathbf{U},\mathbf{V}) = 0 \;\; \text{and} \\
 g(\mathbf{V}|\mathbf{U}_{\ast},\mathbf{V}) - g(\mathbf{V}_{\ast}| \mathbf{U}_{\ast},\mathbf{V}) = 0. 
\end{align}
Since both  $l(\mathbf{U}|\mathbf{U},\mathbf{V})$ and $ g(\mathbf{V}|\mathbf{U}_{\ast},\mathbf{V})$ are strictly convex functions, $\mathbf{U}_{\ast}$ and $\mathbf{V}_{\ast}$ are uniquely acquired. Hence the above equalities hold only if $(\mathbf{U},\mathbf{V})=(\mathbf{U}_{\ast},\mathbf{V}_{\ast})$, that is $(\mathbf{U},\mathbf{V})$ is a fixed point of Algorithms 1 and 2. The same procedure can be followed for proving the second argument of the Lemma concerning Algorithm 3. 

\subsection*{Proof of Proposition 3}
From (\ref{lemma_2_a}) by adding $K$ successive terms we get,
\begin{align}
 \sum^K_{k=1} \delta^a_k \leq f(\mathbf{U}_1,\mathbf{V}_1) - f(\mathbf{U}_K,\mathbf{V}_K) \leq f(\mathbf{U}_1,\mathbf{V}_1) - f^{\infty} < \infty \label{prop_3_sum}
\end{align}
Therefore, the sequence $\delta^a_k$ is bounded and hence it contains convergent subsequences. Moreover it can be shown that as $k\rightarrow \infty$, $\underset{1\leq k \leq K}{\mathrm{min}} \delta_k^a  \rightarrow 0$. Hence by Lemma 3 we know that the limit points of $\delta^a_k$ are in fact fixed points of Algorithms 1 and 2. By (\ref{lemma_2_a}) and as a consequence of the continuity of the cost functions, it can be easily seen that these fixed points actually correspond to stationary points thereofs. The rates of convergence arise by substituting the fist part of inequality (\ref{prop_3_sum}) by $K \underset{1\leq k \leq K}{\mathrm{min}} \delta_k^a \leq \sum^K_{k=1} \delta^a_k $. The proof is exactly the same for Algorithm 3, using $\delta^b_k$ in place of $\delta^a_k$.




\end{document}